%% file: main.tex
\documentclass{article} 
\usepackage{iclr2026_conference,times}

\iclrfinalcopy

\usepackage{xcolor}
\usepackage{colortbl}
\definecolor{customblue}{RGB}{135,200,235}
\definecolor{customred}{RGB}{255,204,204}
\newcommand{\bluebg}[1]{\colorbox{customblue!30}{#1}}
\newcommand{\redbg}[1]{\colorbox{customred!30}{#1}}

\input{math_commands.tex}

\usepackage{wrapfig}
\usepackage[hidelinks]{hyperref}
\usepackage{url}
\usepackage{booktabs}
\usepackage{graphicx}
\usepackage{subfigure}
\usepackage{algorithm}
\usepackage{algpseudocode}
\usepackage{amssymb}
\usepackage{multirow}
\usepackage{makecell}

\title{TyphoonMLA: A Mixed Naive-Absorb MLA Kernel For Shared Prefix}

\author{Ahmet Caner Y\"{u}z\"{u}g\"{u}ler, Ahmet \c{C}elik, Jiawei Zhuang \& Lukas Cavigelli \\
Huawei\\
Switzerland \\
\texttt{\{ahmet.yuzuguler,ahmet.celik,zhuangjiawei,lukas.cavigelli\}@huawei.com} \\
}

\begin{document}

\maketitle

\begin{abstract}
Multi-Head Latent Attention (MLA) is a recent attention mechanism adopted in state-of-the-art LLMs such as DeepSeek-v3 and Kimi K2. Thanks to its novel formulation, MLA allows two functionally equivalent but computationally distinct kernel implementations: naive and absorb. While the naive kernels (e.g., FlashAttention) are typically preferred in training and prefill for their computational efficiency, existing decoding kernels (e.g., FlashMLA) rely on the absorb method to minimize HBM bandwidth usage. However, the compute-bound nature of the absorb implementations prohibits performance benefits from data reuse opportunities in attention calculations, such as shared prefixes. In this work, we introduce TyphoonMLA, a hybrid approach that combines naive and absorb formulations to harness the strengths of both. TyphoonMLA effectively leverages the shared prefix by applying the naive formulation to the compute-bound parts of attention calculations, while reducing the bandwidth requirements for non-shared parts by using the absorb formulation. As a result, TyphoonMLA improves the throughput of attention calculations in MLA architectures by up to 3× and 3.24× on NPU and GPUs, and boosts end-to-end throughput by up to 1.48× in tokens per second, with only a 3\% overhead in HBM size.
\end{abstract}

\input{tex/1_intro}

\input{tex/2_background}

\input{tex/3_proposed_method}

\input{tex/4_experiments}

\input{tex/5_related_work}
\input{tex/6_conclusion}

\newpage

\bibliography{biblio}
\bibliographystyle{iclr2026_conference}

\newpage

\input{tex/9_appendix}

\end{document}

%% file: math_commands.tex
\usepackage{amsmath,amsfonts,bm}

\def\eqref#1{equation~\ref{#1}}

\def\1{\bm{1}}

\DeclareMathAlphabet{\mathsfit}{\encodingdefault}{\sfdefault}{m}{sl}
\SetMathAlphabet{\mathsfit}{bold}{\encodingdefault}{\sfdefault}{bx}{n}



%% file: tex/1_intro.tex
\section{Introduction}

Large Language Models (LLMs) have been widely adopted in various application domains, ranging from chat assistants~\citep{OpenAI23} to coding agents~\citep{Chen21}, thanks to their unprecedented language processing and reasoning capabilities. However, their substantial computational requirements result in slow and inefficient inference, which undermines the user experience and increases operational costs. Therefore, addressing these computational challenges is essential for enabling broader adoption and ensuring their sustainable deployment.

To improve the efficiency of LLM inference, a new attention architecture called Multi-Head Latent Attention (MLA) has recently been introduced \citep{DeepSeek2024}. In this architecture, the key and value tensors that hold contextual information from previous tokens, i.e., the KV-cache, are stored in a low-rank latent space, which helps eliminate memory bandwidth bottlenecks in attention layers. MLA's flexibility to merge the latent-space projection layers with matrix multiplication operations in attention calculations (i.e., \textit{absorption} trick) makes it possible to implement MLA in two functionally equivalent but computationally distinct ways, namely, \textbf{naive} and \textbf{absorb}. 

In training and prefill, where performance is typically limited by the compute capacity (e.g., the total throughput of Tensor cores in GPUs), the compute-efficient naive implementation is preferred. In decode, where the HBM bandwidth dictates the performance, the memory-efficient absorb implementation is used. As a result, MLA architectures utilize computational resources more effectively both in training and inference, making them a more efficient alternative to other attention architectures, such as Multi-Head Attention (MHA) \citep{Vaswani17} or Grouped-Query Attention (GQA) \citep{Ainslie23}. Consequently, MLA serves as the backbone of various state-of-the-art LLMs, such as DeepSeek-v3~\citep{deepseekv3} and Kimi K2~\citep{kimik2}. 

Another promising direction to improve the efficiency of attention kernels is prefix sharing \citep{Juravsky24}. In various application scenarios, a portion of the KV-cache is shared across multiple queries, creating a data reuse opportunity that can help alleviate memory bottlenecks. For instance, many of the inference services today employ a system prompt, which is shared among all user queries to enhance the safety and quality of generated responses. Both official documentation and recent leaks through prompt-injection attacks have revealed that system prompts used in popular inference services reach lengths of tens of thousands of tokens (e.g., Claude-4 has a system prompt of 26k tokens~\citep{system_prompt_repo}). Furthermore, parallel reasoning techniques, such as Tree-of-thought (ToT)~\citep{Yao23} and Graph-of-thought (GoT)~\citep{Besta24}, as well as the speculative decoding techniques~\citep{Wang25-opt}, often result in multiple queries attending to the same part of the KV-cache, boosting the potential for data reuse in attention computation. 

To effectively benefit from this potential data reuse in attention calculations, recent studies have proposed several MHA and GQA kernel implementations~\citep{Yao25, Pan2025, Wang25-flash}. In these designs, the shared parts of the KV-cache are read only once from HBM and reused across multiple queries, reducing HBM accesses and mitigating memory bandwidth bottlenecks. Although these techniques improve the efficiency and performance of MHA and GQA kernels, they are not directly applicable to MLA, since the performance of a typical MLA kernel implementation is limited by computation rather than memory bandwidth. As a result, current MLA kernels fail to fully exploit the data reuse opportunities available in attention computation.

In this paper, we propose a novel MLA method that effectively leverages data reuse in the shared parts of the KV-cache to improve the efficiency and performance of attention computation. Our key observation that leads to the proposed solution lies in the fact that, while the absorb implementation is preferred in memory-bound regions, the naive implementation becomes more efficient in the compute-bound regions when the KV-cache is shared across multiple queries, as it requires fewer floating-point operations than the absorb implementation. Building on this insight, our proposed MLA method merges the naive and absorb implementations: it uses the naive formulation in the shared parts of the KV-cache to exploit its computational efficiency and the absorb formulation in the non-shared parts of the KV-cache to benefit from its memory bandwidth efficiency. 

We demonstrate through extensive experiments on GPUs and NPUs that TyphoonMLA substantially improves resource utilization and offers a speedup of up to 3.2$\times$. Moreover, the proposed method is compatible with other optimization techniques, such as PagedAttention~\citep{Kwon23} and RadixAttention~\citep{Zheng24}, and it supports existing parallelization strategies, such as tensor and sequence parallelism. Therefore, TyphoonMLA can be effortlessly integrated into popular inference frameworks, such as vLLM and SGLang. Furthermore, the proposed method is mathematically equivalent to standard MLA implementations; thus, it does not cause any accuracy degradation and requires neither training nor fine-tuning.

In short, this paper makes the following contributions:
\begin{itemize}
    \item We introduce TyphoonMLA, a novel MLA inference method that achieves higher computational efficiency and resource utilization in attention calculations. To the best of our knowledge, this is the first work that combines both naive and absorb implementations in MLA calculations.
    
    \item We provide a detailed analysis that shows that TyphoonMLA requires fewer floating-point operations in compute-bound regions and consumes less HBM bandwidth in the memory-bound regions than existing MLA kernels.
    
    \item We conduct a series of experiments on NPUs and GPUs, demonstrating that TyphoonMLA improves attention throughput by up to 3.2$\times$ for DeepSeek-v3 and Kimi K2, with only a 3\% overhead in memory footprint.

\end{itemize}

Our code is open-sourced and publicly available\footnote{https://github.com/huawei-csl/TyphoonMLA-community}. Generative AI tools were used to edit and refine this paper to improve the clarity and quality of writing.

%% file: tex/2_background.tex
\begin{figure}[ht!]
    \centering
    \includegraphics[width=1.\linewidth]{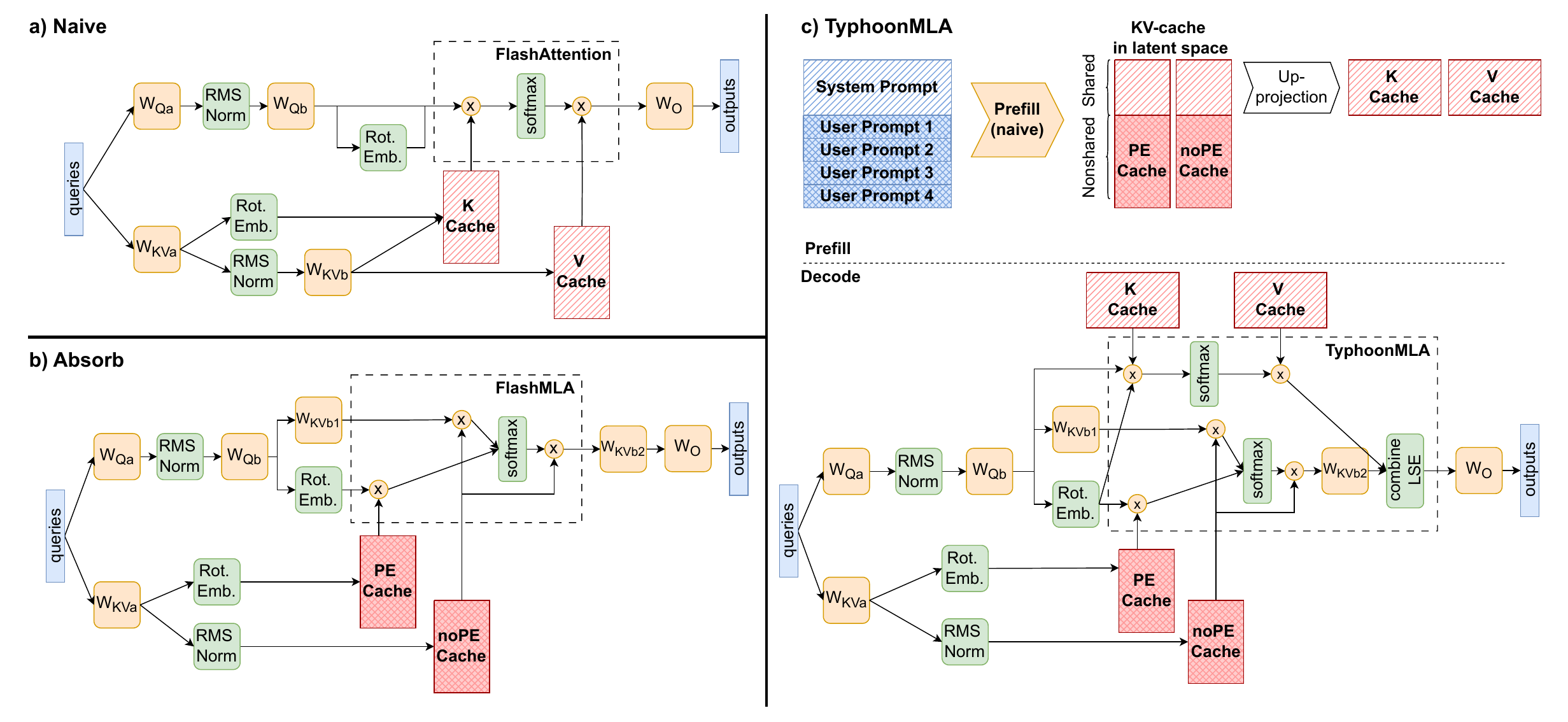}
    \caption{a) The naive formulation of MLA. b) The absorb formulation of MLA. c) The prefill and decode stages of TyphoonMLA.}
    \label{fig:method-overview}
\end{figure}

\section{Background}
In this section, we first provide background on MLA and describe its two implementation variants: naive and absorb. We then discuss prefix sharing in LLM inference and its potential for improving the efficiency of self-attention kernels in MLA layers. 

\subsection{Multi-Head Latent Attention}
MLA is an attention mechanism introduced by DeepSeek-v2 and adopted by state-of-the-art LLMs such as DeepSeek-v3~\citep{deepseekv3}, Kimi K2~\citep{kimik2}, and LongCat~\citep{longcat25}. Unlike standard attention mechanisms, MLA employs a low-rank key-value joint compression mechanism to reduce KV-cache size and bandwidth requirements during inference. Moreover, its novel positional encoding enables the key-value projections to be rearranged in a way that leads to two computationally distinct implementations, namely naive and absorb.

\textbf{Naive:} 
The naive implementation of MLA keeps the KV-cache in an uncompressed form. Fig.~\ref{fig:method-overview}(a) illustrates the components of  MLA in a naive implementation. Therein, after passing through the rotational embedding and RMS normalization layers, the key and value tensors are decompressed using the up-projection matrix, denoted as $W_{KVb}$ in the figure. As a result, the K and V caches consist of multiple attention heads, and the self-attention calculation becomes equivalent to the standard MHA formulation. 

\textbf{Absorb:} 
The absorb implementation of MLA utilizes the commutative property of matrix multiplication to reposition the up-scaling matrix. In absorb formulation, as shown in Fig.~\ref{fig:method-overview}(b), the up-projection matrix is split into two submatrices (denoted as $W_{KVb1}$ and $W_{KVb2}$) and placed before and after the query ($W_{Qb}$) and output ($W_{O}$) projection layers. This formulation allows the KV-cache to remain in a compressed form, represented as PE and noPE caches in the figure. However, applying the up-projection matrix to the queries further expands their embedding dimensions, requiring more floating-point operations in the self-attention calculations. Moreover, the resulting formulation is incompatible with existing self-attention kernels (e.g., FlashAttention~\citep{Dao22}), necessitating the development of specialized kernels for MLA layers (e.g., FlashMLA~\citep{flashmla2025}, ThunderMLA~\citep{Spector25}).

\subsection{Prefix Sharing}
In various inference scenarios, large portions of the KV-cache are shared across multiple queries, creating opportunities for data reuse. First, modern inference services often employ system prompts to enhance the safety and quality of generated responses. These system prompts include tool usage instructions, API call descriptions, and even recent news updates after the model's knowledge cutoff dates, and can consist of tens of thousands of tokens~\citep{system_prompt_repo}. Second, recent LLM reasoning techniques, such as Chain-of-Thought with self-consistency~\citep{Wang23}, Tree-of-Thought~\citep{Yao23}, and Graph-of-Thought~\citep{Besta24}, reformulate inference as a tree or graph search, in which multiple branches run in parallel while sharing a common prefix. Finally, certain speculative decoding techniques~\citep{Wang25-opt} require validating multiple candidate tokens in parallel, which share a long sequence of past tokens. In all these cases, multiple queries attend to overlapping regions of the KV-cache, offering a significant data reuse opportunity to improve the efficiency and performance of attention computations.

%% file: tex/3_proposed_method.tex
\section{Proposed Method}
In the previous section, we discussed two distinct ways of implementing MLA. In this section, we provide details about TyphoonMLA, which combines both implementations.

\subsection{TyphoonMLA}
As previously discussed, the shared prefix introduces data reuse in the shared parts of the attention calculations. Building on this insight, TyphoonMLA partitions the attention calculation into low- and high-arithmetic-intensity components and applies absorb and naive formulations, respectively. Fig.~\ref{fig:method-overview}(c) illustrates how TyphoonMLA works. 

\textbf{Prefill:} 
In the prefill stage, a tree-like query structure is formed from the incoming user requests and a shared prefix (e.g., a system prompt). The queries are processed by the LLM using a prefix-aware naive kernel, producing PE and noPE caches in the low-rank latent space. The portion of the PE and noPE caches that corresponds to the shared prefix is then expanded via the up-projection to form the K and V caches. Consequently, the shared and non-shared parts of the KV-cache are stored in uncompressed and compressed formats, respectively, enabling the use of both naive and absorb formulations. Notably, the up-projection incurs no additional computational overhead, as it is already performed by standard naive kernels during the prefill stage.

\begin{algorithm}[b]

\caption{TyphoonMLA Decode Attention Kernel} \label{alg:typhoonMLA}
\small
\begin{algorithmic}[1] 
    \Require Queries after $\mathbf{W}_{Qb}$ proj. and RoPE layers, $\mathbf{Q} \in \mathbb{R}^{B \times H \times D_{qk}}$ 
    \Require Shared KV-cache in uncompressed form, $\mathbf{C}_{K} \in \mathbb{R}^{B \times L_s \times D_{qk}}$, $\mathbf{C}_{V} \in \mathbb{R}^{B \times L_s \times D_{v}}$
    \Require Non-shared KV-cache in latent space, $\mathbf{C}_{N} \in \mathbb{R}^{B \times L_n \times D_{l}}$, $\mathbf{C}_{R}  \in \mathbb{R}^{B \times L_n \times D_{r}}$
    \Require KV up-projection matrices, $\mathbf{W}_{KVb1} \in \mathbb{R}^{H \times D_n \times D_{l}}$, $\mathbf{W}_{KVb2}  \in \mathbb{R}^{H \times D_v \times D_{l}}$
    
    \Statex 

    \State $[\mathbf{Q}_{N}, \mathbf{Q}_{R}] \gets \mathbf{Q}$ \Comment{Split $\mathbf{Q}$ from dimension $D_{qk}$ into $[D_n,D_r]$}
    \State $\mathbf{Q}_{R} \gets \text{RoPE}(\mathbf{Q}_{R})$ \Comment{Apply positional encoding}
    \State $\mathbf{Q}_{K} \gets [\mathbf{Q}_{N}, \mathbf{Q}_{R}]$ \Comment{Merge after RoPE}
    \State $\mathbf{O}_N, l_{N} \gets \text{softmax}(\mathbf{Q}_{K}{\mathbf{C}_{K}}^{\intercal})\mathbf{C}_{V}$ \Comment{Compute naive component}
    \State $\mathbf{Q}_{A} \gets \mathbf{Q}_{N}\mathbf{W}_{KVb1}$
    \State $\mathbf{O}_A, l_{A} \gets \text{softmax}(\mathbf{Q}_{A}\mathbf{C}_{N}^{\intercal}+\mathbf{Q}_{R}\mathbf{C}_{R}^{\intercal})\mathbf{C}_{N}$ \Comment{Compute absorb component}
    \State $\mathbf{O}_A \gets \mathbf{O}_A \mathbf{W}_{KVb2}^{\intercal}$
    \State $\mathbf{O} \gets \text{combine}(\mathbf{O}_N, \mathbf{O}_A, l_{N}, l_{A})$ \Comment{Combine partial outputs using LSEs}
    \State \textbf{return} $\mathbf{O}$ 
    
\end{algorithmic}
\end{algorithm}

\textbf{Decode:} In the decode stage, TyphoonMLA first applies the down-projection layers $W_{Qa}$ and $W_{KVa}$ to the queries, obtaining the \textit{q} and \textit{kv} vectors. The \textit{q} vectors pass through an RMS normalization layer, followed by the up-projection layer $W_{Qb}$. The \textit{kv} vectors are split into two subvectors, one of which passes through a rotational embedding layer, and the other through an RMS normalization. All operations up to this point are common to TyphoonMLA, naive, and absorb implementations.

Unlike naive and absorb implementations, TyphoonMLA performs self-attention calculations using both compressed and uncompressed KV-cache while reusing the shared operations. As described in Algorithm \ref{alg:typhoonMLA}, TyphoonMLA takes the output of $W_{Qb}$ up-projection layer ($Q$), shared KV-cache in uncompressed form ($C_K$ and $C_V$), non-shared KV-cache in latent form ($C_N$ and $C_R$), and KV up-projection matrices ($W_{KVb1}$ and $W_{KVb2}$) as inputs. The input queries $Q$ are first split from their embedding dimension into two tensors, and a positional encoding is applied. Then, similar to the naive implementation, the queries are multiplied by the uncompressed K cache, $C_K$, passed through a softmax function, and multiplied by the V cache, $C_V$. Similar to the absorb implementation, the queries are first up-projected using the $W_{KVb1}$ matrix, multiplied by the compressed PE and noPE caches ($C_R$ and $C_N$), passed through a softmax function, and multiplied again by the noPE cache. As required by the absorb formulation, the output of the absorb component is then up-projected once again using the $W_{KVb2}$ matrix. Finally, the partial results from these two components are then aggregated using a \textit{combine} function with the help of the log-sum-exp (LSE) of the softmax denominators ($l_N$ and $l_A$). TyphoonMLA is mathematically equivalent to both absorb and naive implementations; therefore, it does not require any re-training or fine-tuning.

\textbf{Fall-back to Absorb:} At small batch sizes, when there is insufficient data reuse, the naive implementation can become less efficient than absorb due to higher memory access costs. To address this, TyphoonMLA automatically switches to an absorb-only kernel whenever the batch size falls below a predefined threshold. By doing so, TyphoonMLA avoids any performance penalty at small batch sizes, ensuring consistently high efficiency across a wide range of batch sizes.

\textbf{Parallelization:} TyphoonMLA is fully compatible with existing parallelization strategies for attention computation. Although the compressed PE and noPE caches have a single head dimension, the uncompressed K and V caches can be parallelized across attention heads. Additionally, both compressed and uncompressed caches can be easily parallelized across the sequence dimension. As a result, TyphoonMLA seamlessly supports both tensor and sequence parallelism, enabling efficient scaling across multiple NPUs or GPUs.

\begin{table}[t!]
\centering
\small
\caption{Computational analysis of naive, absorb, and TyphoonMLA. $B$: batch size, $S_q$: query sequence length, $L_{s}$: shared context length, $L_{n}$: non-shared context length, $H$: number of heads, $D_{qk}$: head dim. for Q and K, $D_{v}$: head dim for V, $D_l$: KV LoRA rank, $D_n$: noPE head dim, $D_r$: RoPE head dim. TyphoonMLA always requires smaller memory operations than naive (highlighted in \redbg{red}) and fewer multiply-accumulate operations (MACs) than absorb (highlighted in \bluebg{blue}).}
\label{table:complexity}
\begin{center}
\begin{tabular}{rll}
\toprule
& \bf \textsc{MAC} & \textsc{DeepSeek-v3 \scriptsize{($\times 1024$)}}\\
\cmidrule(r){2-2}\cmidrule(l){3-3}
\bf \textsc{Naive} & $BS_qL_sH(D_{qk}+D_v) + BS_qL_nH(D_{qk}+D_v)$ & $40 \times BL_s+40 \times BL_n$ \\
\bf \textsc{Absorb}  & $BS_qL_sH(2D_l+D_r) + BS_qL_nH(2D_l+D_r)$ & \bluebg{$136 \times BL_s$}  $+$ $136 \times BL_n$ \\
\bf \textsc{TyphoonMLA}  & $BS_qL_sH(D_{qk}+D_v) + BS_qL_nH(2D_l+D_r)$ & \bluebg{$40 \times BL_s$} $ + 136 \times BL_n$ \\
\midrule
& \bf \textsc{\textsc{HBM R/W}} & \textsc{DeepSeek-v3 \scriptsize{($\times 1024$)}} \\
\cmidrule(r){2-2}\cmidrule(l){3-3}
\bf \textsc{Naive} & $L_sH(D_{qk}+D_v) + BL_nH(D_{qk}+D_v)$ & $40 \times L_s$ $+$  \redbg{$40 \times BL_n$} \\
\bf \textsc{Absorb} & $L_s(D_l + D_r) + BL_n(D_l + D_r)$ & $0.56 \times L_s+ 0.56 \times BL_n$ \\
\bf \textsc{TyphoonMLA} & $L_sH(D_{qk}+D_v) + BL_n(D_l + D_r)$ & $40 \times L_s$ $+$ \redbg{$0.56 \times BL_n$} \\
\bottomrule
\end{tabular}
\end{center}
\label{tab:complexity}
\end{table}

\subsection{Computational Analysis}
\label{sec:complexity}
To quantify the potential performance benefits of the proposed method, we derive the computational requirements of attention calculations in terms of the number of multiply-accumulate operations (MAC) and the size of memory read/write operations (HBM R/W) for TyphoonMLA, as well as the naive and absorb formulations. Table \ref{tab:complexity} summarizes these derivations with respect to architectural and generation parameters. In the right-most column, we substitute the architectural parameters with those of the DeepSeek-v3 to facilitate a direct comparison. In our analysis, we consider only self-attention computations, excluding the projection layers, since self-attention dominates execution time at long sequence lengths and asymptotically approaches the total runtime.

Since the absorb implementation is typically compute-bound, its performance is primarily determined by the number of MACs. In contrast, the naive implementation is generally memory-bound, so its efficiency is dominated by the amount of data read from HBM. Therefore, to maximize overall performance, TyphoonMLA aims to achieve lower MACs than absorb in compute-bound regions and fewer HBM read and writes than naive in memory-bound regions.

Table \ref{tab:complexity} shows that the naive implementation requires reading $(40 \times L_s+40 \times BL_n) \times 1024$ words from HBM. In comparison, TyphoonMLA reads only $(40 \times L_s + 0.56 \times BL_n)\times1024$ words, which is the same for the shared portion but approximately 70× smaller for the non-shared portion. The absorb implementation, typically compute-bound, requires $(136 \times BL_s + 136 \times BL_n)\times 1024$ MAC operations. TyphoonMLA, in contrast, requires only $(40 \times BL_s + 136 \times BL_n) \times 1024$ MAC operations, which is the same for the non-shared portion, but 3.4× smaller in the shared portion. This analysis reveals that TyphoonMLA requires \textbf{fewer bytes} to read from HBM than the naive formulation in memory-bound regions and \textbf{fewer MACs} than the absorb formulation in compute-bound regions, reducing the overall computational complexity of attention calculations in MLA.

To combine the partial results of the naive and absorb parts, TyphoonMLA uses a \textit{CombineLSE} function, similar to the epilogue stage in Flash Attention \citep{Dao22}. This function performs only vector operations and requires reading $2BS_qHD_v$ bytes from HBM and performing $2BS_qHD_v$ MAC operations. Since the computational complexity of this function is independent of the KV sequence length, which typically ranges from hundreds to thousands, its computational overhead is insignificant compared to the other components of the attention calculations.

For TyphoonMLA to achieve a speedup, reading the shared portion of the KV-cache should be faster with the naive implementation than computation with the absorb implementation. This happens only when there is sufficient data reuse, which occurs when the batch size is larger than a threshold. By equating the memory read time of the naive and the computation time of the absorb implementation for the shared part of the KV-cache, we identify the batch size threshold, $B_\theta$, in terms of architectural parameters (defined above) and hardware specifications, namely throughput $T$ and HBM memory bandwidth $M$, as follows:

\begin{equation}
    \frac{L_sH(D_{qk}+D_v)}{M} = \frac{{B_\theta}S_qL_sH(2D_l+D_r)}{T} \implies
    B_\theta = \frac{(D_{qk}+D_v)}{S_q(2D_l+D_r)} \frac{T}{M}
\end{equation}

When the architectural parameters are replaced with those of DeepSeek-v3 and the hardware parameters with Ascend NPU ($T=376$~TOPS/s, $M=1.8$~TB/s), we obtain $B_{\theta} = 61$. This threshold indicates the break-even point, beyond which TyphoonMLA becomes computationally more efficient than the absorb implementation, whereas, for batch sizes smaller than $B_\theta$, TyphoonMLA falls back to the absorb kernel to avoid any slowdown.

%% file: tex/4_experiments.tex
\begin{figure}
    \centering
    \includegraphics[width=1\linewidth]{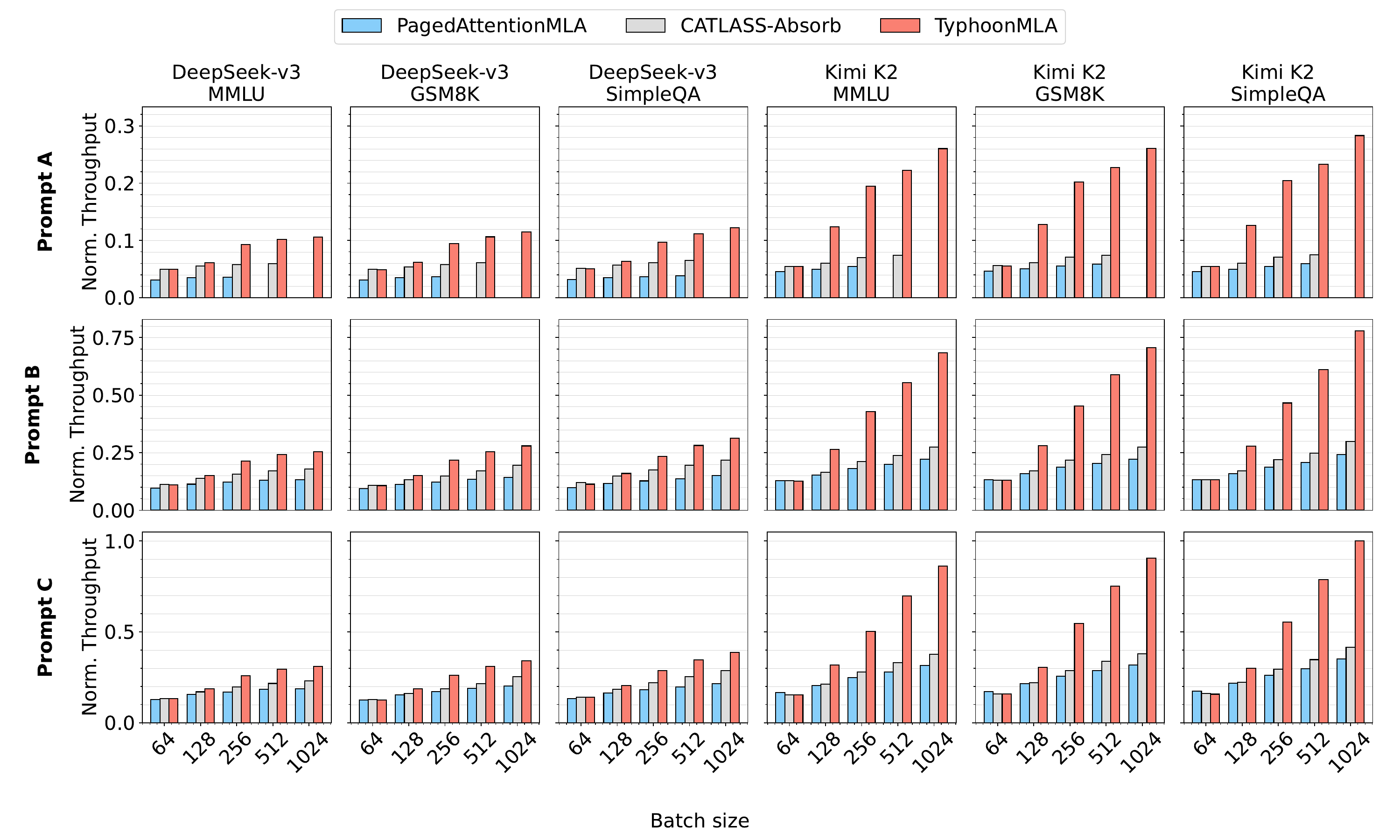}
    \caption{Benchmark results on Ascend NPUs. Y-axes represent normalized throughput in terms of the number of generated tokens per second per layer. Some data points for baselines are missing as their memory footprint exceeds the HBM capacity.}
    \label{fig:main-results-ascend}
\end{figure}

\section{Experiments}
Building on the complexity analysis presented in Section~\ref{sec:complexity}, we now provide empirical evidence to validate the performance benefits of TyphoonMLA. In this section, we evaluate the performance of TyphoonMLA and compare it against various baselines on NPUs and GPUs. First, we measure TyphoonMLA’s throughput on popular datasets using DeepSeek-v3 and Kimi K2 with different system prompts. Next, we examine its performance breakdown to validate our insights. Finally, we assess TyphoonMLA's memory footprint and compare it against the absorb baseline.

\textbf{Experiments on NPUs:}
We implemented TyphoonMLA for Ascend NPUs using the CANN toolkit~\citep{cann}. For the absorb component of TyphoonMLA, we developed a custom kernel that supports paged and variable-length KV-cache using the Ascend CATLASS library~\citep{catlass}. For the naive component of TyphoonMLA, we employed the \texttt{NpuFusedInferAttentionScore} function from TorchNPU~\citep{torchnpu}. For the projection layers and combine logic, we used TorchNPU's \texttt{einsum} and vector operators. 

To evaluate the impact of system prompt length on performance, we selected three system prompts with varying lengths, summarized in Table \ref{tab:prompts}. As benchmark datasets, we used the MMLU~\citep{Hendrycks21}, GSM8K~\citep{Cobbe21}, and SimpleQA~\citep{Wei24}, which comprise questions and answers from various topics. We repeat each experiment for batch sizes of 64, 128, 256, 512, and 1024. 

To simulate a realistic deployment scenario, we adopted continuous batching with a paged KV-cache with a block size of 128. Each experiment starts by randomly sampling questions from a dataset and forming a batch of queries. At the end of each decode iteration, the completed queries are replaced with new questions sampled from the dataset. Each experiment is continued until the entire dataset is processed. The throughput is then calculated by dividing the total number of generated tokens by the total execution time across all decode iterations. We calculate the speedup as the throughput of TyphoonMLA divided by the best of the two baselines. 

\begin{wraptable}{r}{0.4\textwidth} 
\centering
\caption{System prompts used in the experiments, taken from \citep{system_prompt_repo}.}
\label{tab:prompts}
\begin{tabular}{ccr}
\toprule
Name & LLM service & \#tokens \\
\midrule
Prompt A & Claude-4 & 26472 \\
Prompt B & OpenAI/o3 & 7069 \\
Prompt C & Grok/Personas & 4759 \\
\bottomrule
\end{tabular}
\end{wraptable}

We run the experiments on an Ascend NPU, equipped with 24 Davinci cores and 64 GB of HBM memory, which provides a throughput of 376 TOPS/s in FP16 precision and an HBM bandwidth of 1.8TB/s. We compared TyphoonMLA's throughput with that of the baseline kernels, namely TorchNPU PagedAttentionMLA kernel and a custom-built CATLASS Absorb-only kernel. 

Fig.~\ref{fig:main-results-ascend} presents the normalized throughput of TyphoonMLA and the baseline methods for the attention layers of DeepSeek-v3 and Kimi K2. TyphoonMLA consistently outperforms both baselines across all datasets, system prompts, and batch sizes, achieving speedups between 1.2× and 3×. As expected, TyphoonMLA offers the highest speedup with Prompt A, since longer system prompts increase the ratio of shared to non-shared portions of the KV-cache. We further observe that the speedup is generally higher for the Kimi K2 than for DeepSeek-v3. This is because the number of attention heads in Kimi K2 is 64, half of DeepSeek-v3, which makes the former's performance more sensitive to data reuse. Overall, these results confirm that TyphoonMLA effectively exploits the shared prefix and delivers consistent and significant speedups over the existing MLA kernels.

\begin{figure}
    \centering
    \includegraphics[width=1\linewidth]{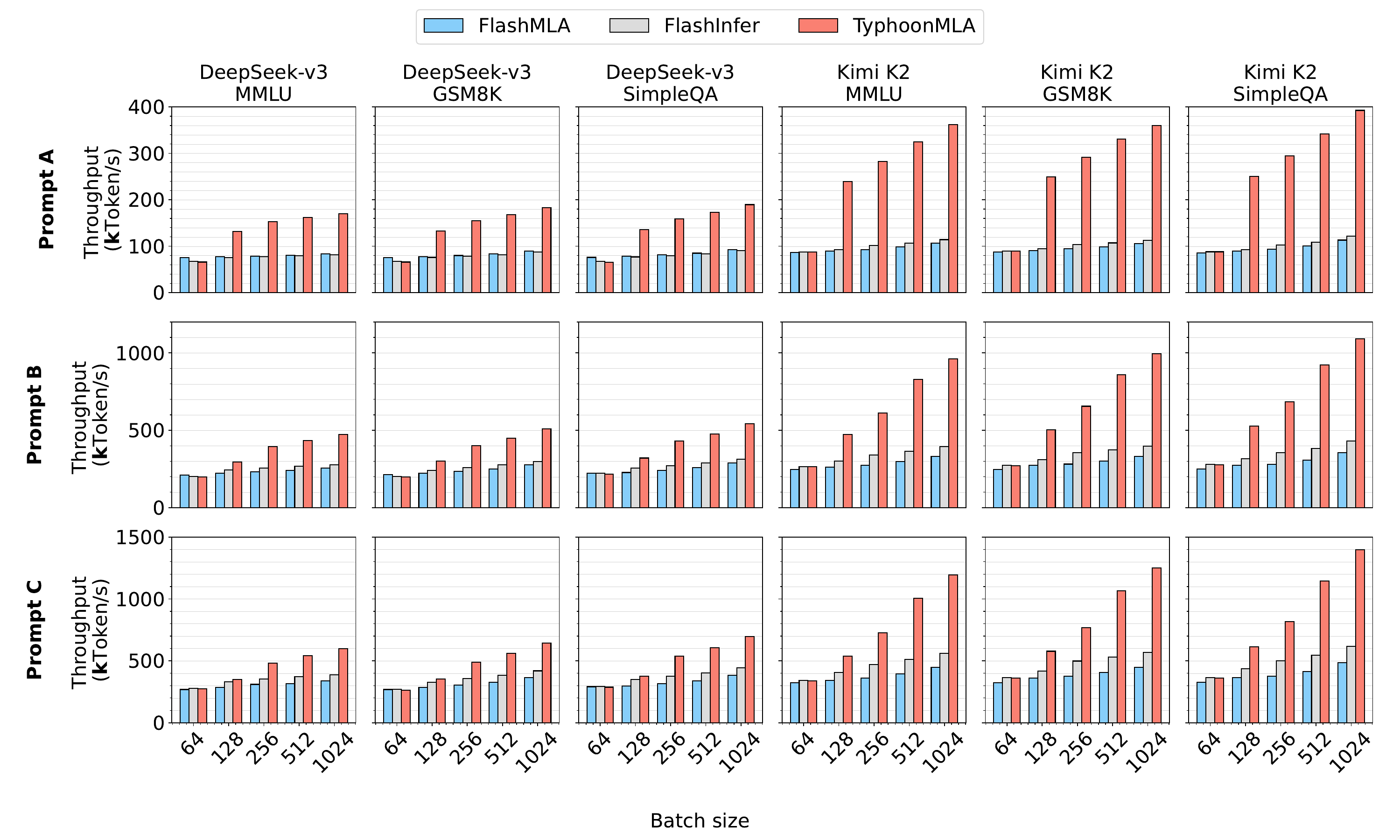}
    \caption{Benchmark results on GPU for various batch sizes. Y-axes represent throughput in terms of the number of generated tokens per second per layer.}
    \label{fig:main-results-gpu}
\end{figure}

\textbf{Experiments on GPUs:} We also implemented TyphoonMLA using the FlashInfer naive and absorb MLA kernels~\citep{ye25} and repeated our NPU experiments on a GPU. Fig.~\ref{fig:main-results-gpu} reports the normalized throughput of TyphoonMLA on a GPU with a 1 PetaFLOPS/s of theoretical throughput in FP16, 3.3 TB/s of HBM bandwidth, comparing it against the FlashMLA~\citep{flashmla2025} and FlashInfer (absorb-only) baselines at batch sizes of 64, 128, 256, 512, and 1024. TyphoonMLA achieves higher throughputs than both baselines, with factors up to 3.24×. Similar to the NPU results, TyphoonMLA accelerates Kimi K2 more than DeepSeek-v3, owing to its smaller number of attention heads. Moreover, it achieves the highest speedups with Prompt A, as it has the highest token count. Overall, these results demonstrate that TyphoonMLA generalizes effectively to GPUs, delivering substantial performance gains across diverse hardware platforms.

\begin{table}[t!]
\centering
\small
\centering
\caption{Token generation rate for DeepSeek-v3 processing MMLU dataset with a batch size of 128 per GPU.}
\label{tab:endtoend}
\begin{tabular}{ccccccc}
\toprule
 & \multicolumn{3}{c}{FlashMLA} & \multicolumn{3}{c}{TyphoonMLA}  \\
 \cmidrule(rl){2-4}\cmidrule(rl){5-7}
 & \makecell{Attention time \\ (ms)} & \makecell{Total time \\ (ms)} & \makecell{TGR \\ (\textbf{k}Token/s)} & \makecell{Attention time \\ (ms)} & \makecell{Total time \\ (ms)} & \makecell{TGR \\ (\textbf{k}Token/s)} \\
\midrule
Prompt A & 99.1 & 127.2 & 1.01 & 58.1 & 86.3 & 1.48 \\
Prompt B & 34.5 & 62.6 & 2.04 & 25.9 & 54.0 & 2.37 \\
Prompt C & 26.9 & 55.0 & 2.33 & 22.0 & 50.1 & 2.56 \\
\bottomrule
\end{tabular}
\end{table}

\begin{wrapfigure}{r}{0.4\textwidth}
    \centering
    \includegraphics[width=1\linewidth]{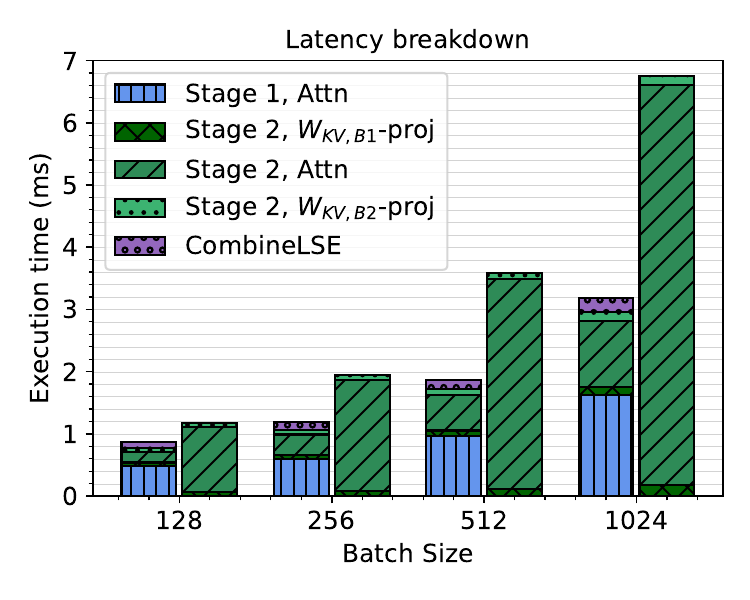}
    \caption{Latency breakdown of TyphoonMLA (bars on the left-hand side) and CATLASS absorb-only baseline (bars on the right-hand side) for Kimi K2 architecture. Stage 1 and Stage 2 represent the naive and absorb parts of TyphoonMLA, and have sequence lengths of 4096 and 512, respectively.}
    \label{fig:breakdown}
\end{wrapfigure}

\textbf{Latency Breakdown:} To demonstrate how TyphoonMLA accelerates attention computations and how its components behave under different batch sizes, we profiled its execution using the CANN toolkit's \texttt{msprof} tool and compared it against the CATLASS absorb-only baseline. Fig.~\ref{fig:breakdown} shows the execution time breakdown of TyphoonMLA (bars on the left-hand side) and the CATLASS baseline (bars on the right-hand side). In the figure, we denote the naive and absorb parts of attention computation in TyphoonMLA as \textit{Stage 1, Attn} and \textit{Stage 2 Attn}, respectively; the up and down-scaling required in the absorb formulation as $W_{KV,B1}$-proj and $W_{KV,B2}$-proj; and the combination logic that merges the intermediate results of both stages as \textit{CombineLSE}. We excluded the other linear layers of the attention block, since they are identical across both methods and negligible at long sequence lengths. In this experiment, we set the shared prefix length to 4096, the non-shared sequence length of each request to 512, and used the architectural parameters of Kimi K2. Batch sizes smaller than 128 are omitted, as in these cases, TyphoonMLA falls back to the absorb-only implementation, becoming identical to the baseline. As the CATLASS baseline is absorb-only, it contains only Stage 2 components. 

The runtime breakdown of TyphoonMLA shows that its performance gains align closely with the estimates from our theoretical analysis in Section~\ref{sec:complexity}. 
At a batch size of 1024, the attention calculations in the CATLASS baseline take 6.43 ms, while the naive and absorb parts of TyphoonMLA take 1.63 ms and 1.06 ms, respectively. Since the non-shared part of attention is identical between TyphoonMLA and the baseline, we estimate the execution time of the shared part in the baseline as $6.43 - 1.06 = 5.37$ ms. The ratio between the execution times of the shared parts in the baseline and TyphoonMLA is therefore $5.37 / 1.63 = 3.3$, which matches the reduction in the number of operations between the naive and absorb formulations derived in Section~\ref{sec:complexity}. These results not only validate our initial insights and theoretical analysis but also demonstrate that TyphoonMLA’s hybrid design translates directly into measurable runtime improvements.

\textbf{End-to-End Speedup:} Previous experimental results demonstrated that TyphoonMLA significantly accelerates the attention layers in MLA architectures. However, the execution time of the other layers in LLMs (e.g., MoE, collective communication etc.) would remain unchanged. Therefore, we now investigate the impact of the proposed method on the end-to-end LLM decoding performance. To that end, we estimate the \textit{token generation rate} (TGR) for DeepSeek-v3 with a batch size of 128 per device on a system with 128 GPUs by measuring attention layers on a GPU and using the profiling data for other layers provided by DeepSeek-AI~\citep{deepseek-profile}.

Table \ref{tab:endtoend} reports the estimated TGR and per-iteration execution time for both the attention layers and the full model when using TyphoonMLA, comparing against those with FlashMLA as the baseline. Since attention computation constitutes a substantial portion of overall runtime, the speedup achieved by TyphoonMLA in the attention layers translates directly into notable end-to-end throughput gains, reaching up to a 1.48× improvement in tokens per second.

\begin{figure}[t!]
    \centering
    \includegraphics[width=1\linewidth]{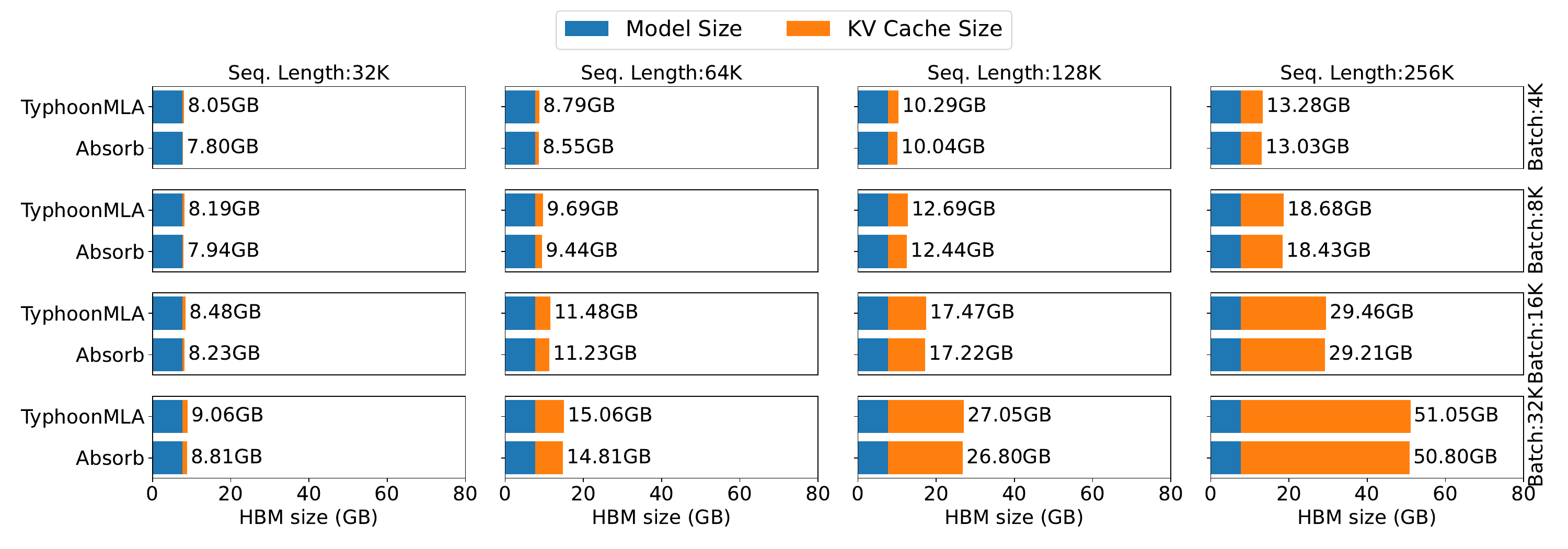}
    \caption{HBM footprint comparison for DeepSeek-v3 in FP8 precision for both weights and KV-cache.}
    \label{fig:hbm-size}
\end{figure}

\textbf{HBM Footprint:} Since TyphoonMLA stores the shared portion of the KV-cache in an uncompressed form, its memory footprint differs from that of the absorb baselines. To quantify the impact of TyphoonMLA on the memory footprint, we analyze TyphoonMLA’s HBM usage under various deployment settings and compare it against the absorb baseline. We assume that the model is distributed across 384 NPUs, as in a CloudMatrix cluster~\citep{Zuo25}, employing full expert parallelism on MoE layers and a combination of data, tensor, and sequence parallelism of factors 24, 4, and 4, respectively. Fig.~\ref{fig:hbm-size} shows the HBM size of DeepSeek-v3 for batch sizes ranging from 4K to 32K and maximum sequence lengths from 32K to 256K, assuming Prompt A as the shared prefix. 

At small batch sizes and sequence lengths, HBM usage is dominated by the model weights, which are identical for both TyphoonMLA and the absorb baseline. As batch size and sequence length increase, the KV-cache grows for both methods. However, at large batch sizes and sequence lengths, the memory occupied by the KV-cache that corresponds to the shared prefix becomes negligible compared to the non-shared portion, which is identical for both methods. Consequently, TyphoonMLA incurs only a minimal HBM overhead, limited to approximately 3\% across a wide range of deployment scenarios.

%% file: tex/5_related_work.tex
\section{Related Work}

\textbf{LLM Serving Acceleration:} Various prior work proposes system-level optimizations for LLM serving. Orca~\citep{Yu22} introduced \textit{continuous batching}, which allows replacing the completed requests in a batch with new ones to improve effective throughput. Inspired by the OS virtual memory concept, PagedAttention~\citep{Kwon23} partitions the KV-cache into memory pages to efficiently handle KV-cache with variable length. TyphoonMLA supports both continuous batching and PagedAttention; thus, it benefits from the performance gains of such system-level optimizations. Furthermore, to distribute LLM inference across multiple devices, existing systems typically employ tensor~\citep{Shoeybi19} and sequence~\citep{Jacobs23} parallelism. TyphoonMLA also seamlessly supports both types of parallelization, enabling deployment at scale.

\textbf{Efficient Attention Kernels:} As attention computation constitutes a significant portion of the total execution time of LLM inference, many prior studies focused on developing efficient attention kernels. Flash Attention~\citep{Dao22} partitions attention computations into tiles that fit on the on-chip buffers to minimize HBM bandwidth usage. FlashDecoding~\citep{dao23} introduces the \textit{Split-K} method to improve the parallelization across GPU cores. Flashinfer~\citep{ye25} provides a customizable attention template to facilitate the development of custom attention kernels. However, all these techniques are developed for MHA; hence, they are not applicable to MLA. FlashMLA~\citep{flashmla2025}, FlashMLA-ETAP~\citep{Dege25}, and ThunderMLA~\citep{Spector25} propose efficient kernel designs for MLA. Unfortunately, these kernels are agnostic to the shared prefix and an absorb-only implementation; therefore, they can not fully exploit hardware resources.

\textbf{Prefix Sharing:} Several prior works (e.g., SGLang, Hydragen etc.) have proposed techniques to exploit shared prefixes in attention calculations and KV-cache management~\citep{Juravsky24,Zheng24}. Other frameworks (e.g., FlashInfer-Cascade, FastTree etc.) developed specialized MHA/GQA GPU kernels for tree-structured KV-caches to leverage the data reuse introduced by shared prefixes~\citep{Ye24, Zhu24, Ye24cascade, Yao25, Pan2025, Wang25-flash}. However, these kernel designs target MHA/GQA architectures and aim to reduce HBM bandwidth usage. Consequently, they are not applicable to MLA, whose decode stage is typically compute-bound. In contrast, TyphoonMLA addresses this gap by reducing the number of FLOPS required in the compute-bound decode stage of MLA, achieving substantial speedups. Moreover, the mechanisms to handle complex tree-structured KV-caches introduced by the prior work would be equally applicable to both naive and absorb implementations. Therefore, applying these techniques to TyphoonMLA would be trivial.

%% file: tex/6_conclusion.tex
\section{Conclusions}
In this paper, we proposed a novel MLA method, TyphoonMLA, which combines the naive and absorb implementations to effectively exploit shared prefixes in the KV-cache. Our theoretical analysis showed that, in the presence of a shared prefix, TyphoonMLA requires fewer FLOPS in compute-bound parts and smaller HBM bandwidth in memory-bound parts of the attention calculations than the existing MLA kernels. Experimental results on NPUs and GPUs show that TyphoonMLA improves attention throughput by up to 3.2× on DeepSeek-v3 and Kimi K2 models, while incurring only a minimal increase in HBM footprint. 

%% file: tex/9_appendix.tex
\appendix
\section{Appendix}

\subsection{Roofline Analysis}

\begin{figure}
    \centering
    \includegraphics[width=1\linewidth]{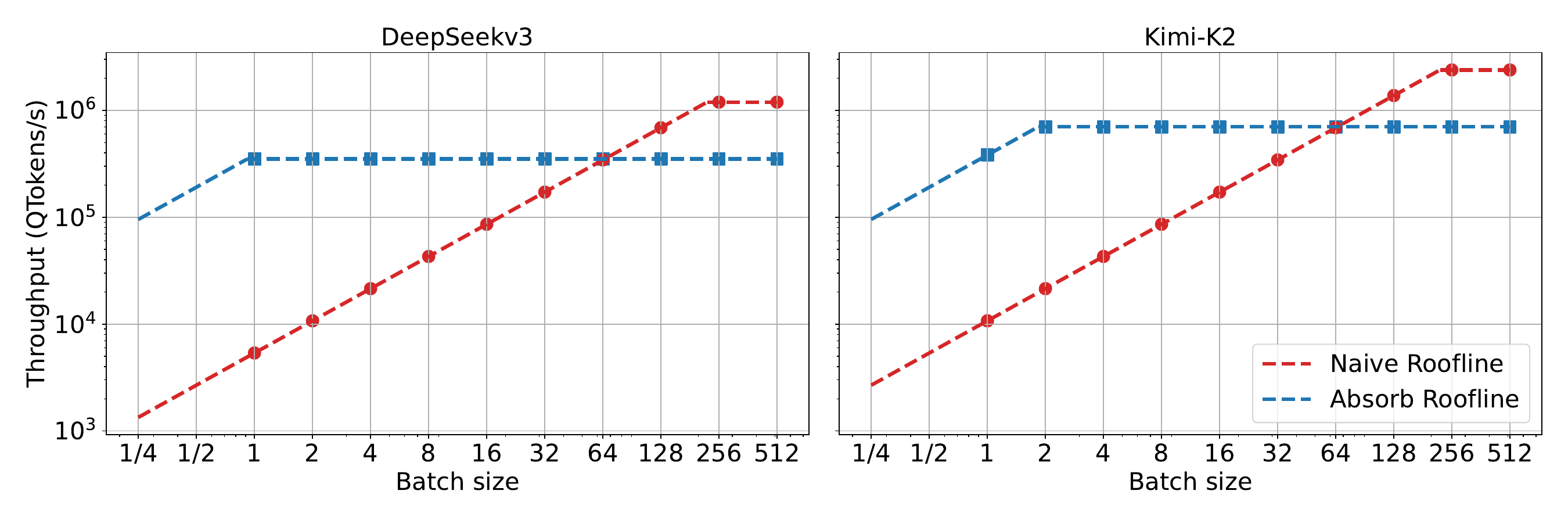}
    \caption{Roofline analysis of the naive and absorb implementations for DeepSeek-v3 and Kimi K2 models with a memory bandwidth of 1.8 TB/s and a cube throughput of 400 TFLOPS/s.}
    \label{fig:roofline}
\end{figure}

To analyze the computational characteristics of the naive and absorb implementations in the presence of a shared prefix, we plot the roofline model~\citep{Williams09} of an AI accelerator for DeepSeek-v3 and Kimi K2 under both implementations, as shown in Fig.\ref{fig:roofline}. The y-axis in the plots represents throughput, calculated as the number of query tokens processed per second by the MLA kernel, while the x-axis represents the batch size, which determines the number of operations per byte (i.e., the operational intensity). The roofline model has two regions of operation: a bandwidth-limited region, where performance is limited by memory bandwidth due to insufficient operational intensity, and a compute-bound region, where performance is limited by the computational capacity of a processor, such as the total throughput of the cube units in NPUs and GPUs.

As expected, the absorb implementation provides higher throughput at low operational intensities thanks to the reduced memory bandwidth usage resulting from the compressed KV-cache stored in latent space. As batch size increases and operational intensity rises, the throughput of the absorb implementation does not improve further. For DeepSeek-v3, performance remains flat once the compute-bound region is reached, while for Kimi K2, throughput quickly saturates beyond a batch size of two. Thus, increasing operational intensity does not improve the performance of absorb implementations, as they are typically compute-bound.

In contrast, the naive implementation greatly benefits from the increasing operational intensity and achieves higher throughputs, as it is typically memory-bound at small batch sizes due to the uncompressed KV-cache. At batch sizes larger than 64, where significant data reuse occurs, the naive implementation achieves up to 3.4$\times$ higher throughput than the absorb implementation, thanks to its lower number of required operations. This analysis indicates that, although the absorb implementation is advantageous at low operational intensities, the naive implementation offers superior performance when sufficient operational intensity is present. These observations motivate our proposed kernel, which combines both approaches to maximize efficiency and performance across both compute and memory-bound regions of MLA calculations.

\subsection{Theoretical Analysis of TyphoonMLA}
\begin{figure}[t!]
    \centering
    \includegraphics[width=1\linewidth]{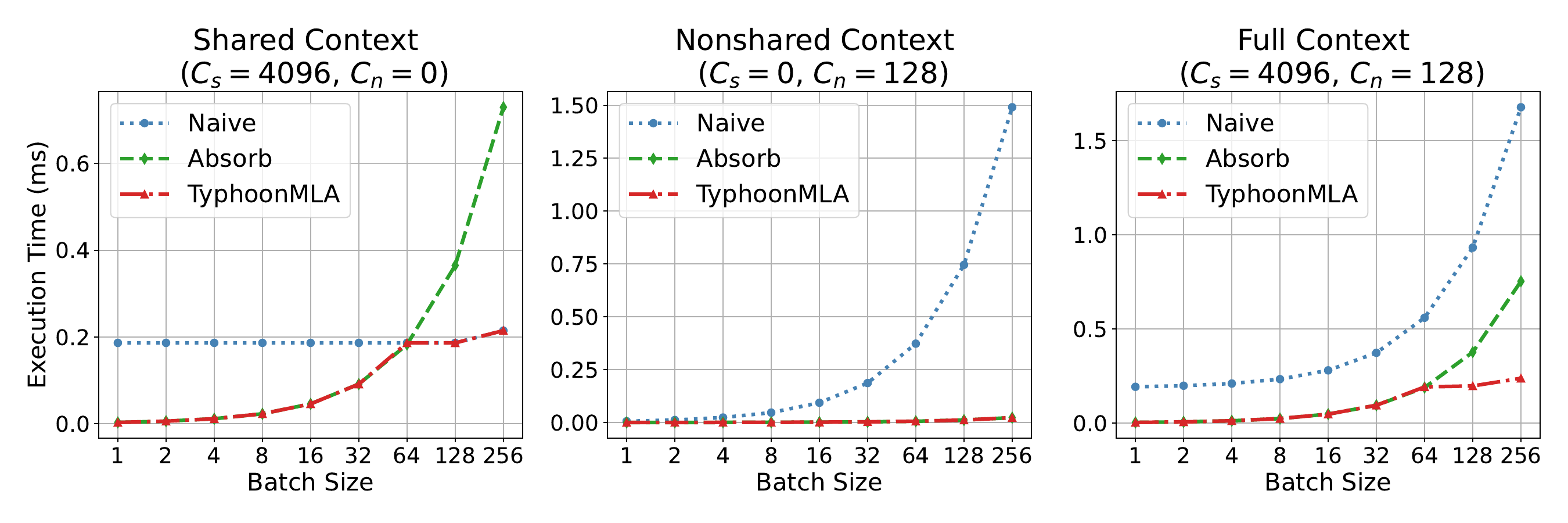}
    \caption{Theoretical analysis of Naive, Absorb, and TyphoonMLA.}
    \label{fig:theoretical}
\end{figure}

In this section, we visualize the computational characteristics of the naive and absorb formulations using the computational model presented in Table~\ref{tab:complexity}. Fig.~\ref{fig:theoretical} depicts the estimated execution time of the absorb implementation, naive implementation, and TyphoonMLA for varying batch sizes.

In the shared context part of the attention calculations, the execution time of the absorb formulation increases linearly with the batch size, since its execution is compute-bound. In contrast, the execution time of the naive formulation remains constant until the batch size of 128, since its execution is memory-bound and the memory bandwidth usage does not change with the batch size in the shared prefix part of the KV-cache. Therefore, although faster at small batch sizes, the absorb becomes slower than the naive at batch sizes larger than 64. Therefore, TyphoonMLA switches from absorb to naive implementation at the batch size of 64 in order to achieve minimum execution time over the full range of batch sizes. 

In the non-shared context part of the attention calculations, the absorb is always faster than naive. Therefore, TyphoonMLA employs the absorb implementation at all batch sizes. As a result of the combination of the two stages, TyphoonMLA is identical to the absorb implementation at batch sizes lower or equal to 64, beyond which the computational benefits of the naive implementation weighs in and makes TyphoonMLA significantly faster than the absorb baseline.

\begin{figure}[b]
\centering     
\subfigure[Shared part of MLA]{\label{fig:benchmark-stage1}\includegraphics[width=0.3\textwidth]{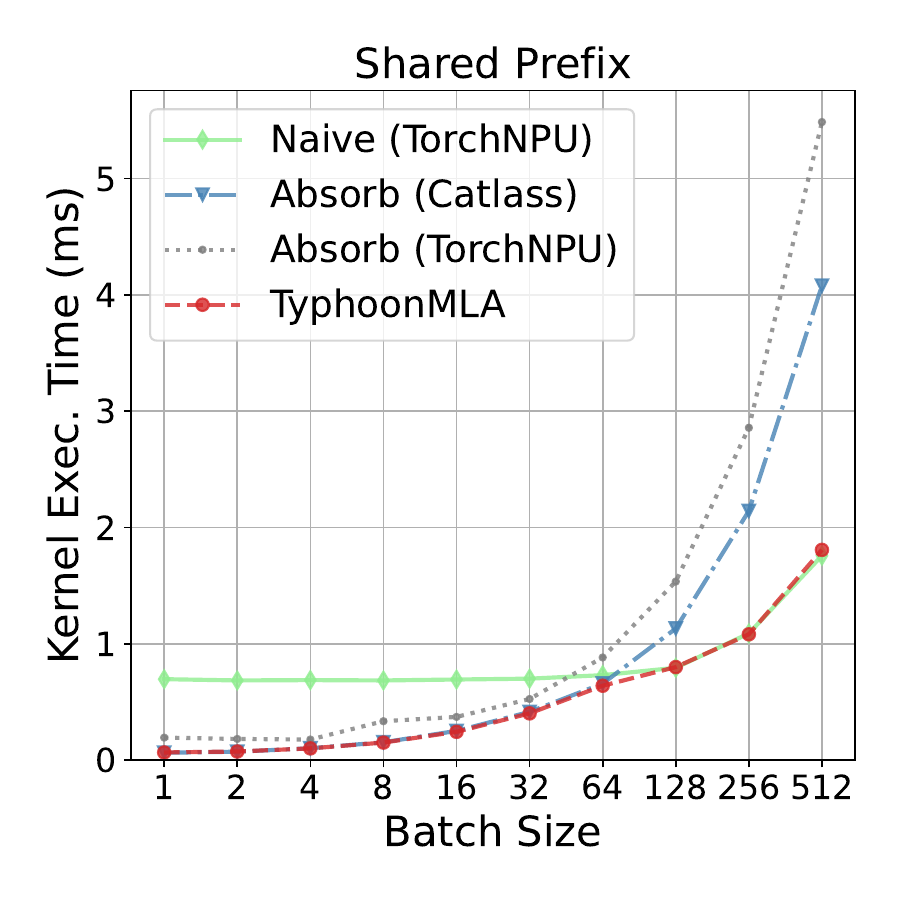}}
\subfigure[Non-shared part of MLA]{\label{fig:benchmark-stage2}\includegraphics[width=0.3\textwidth]{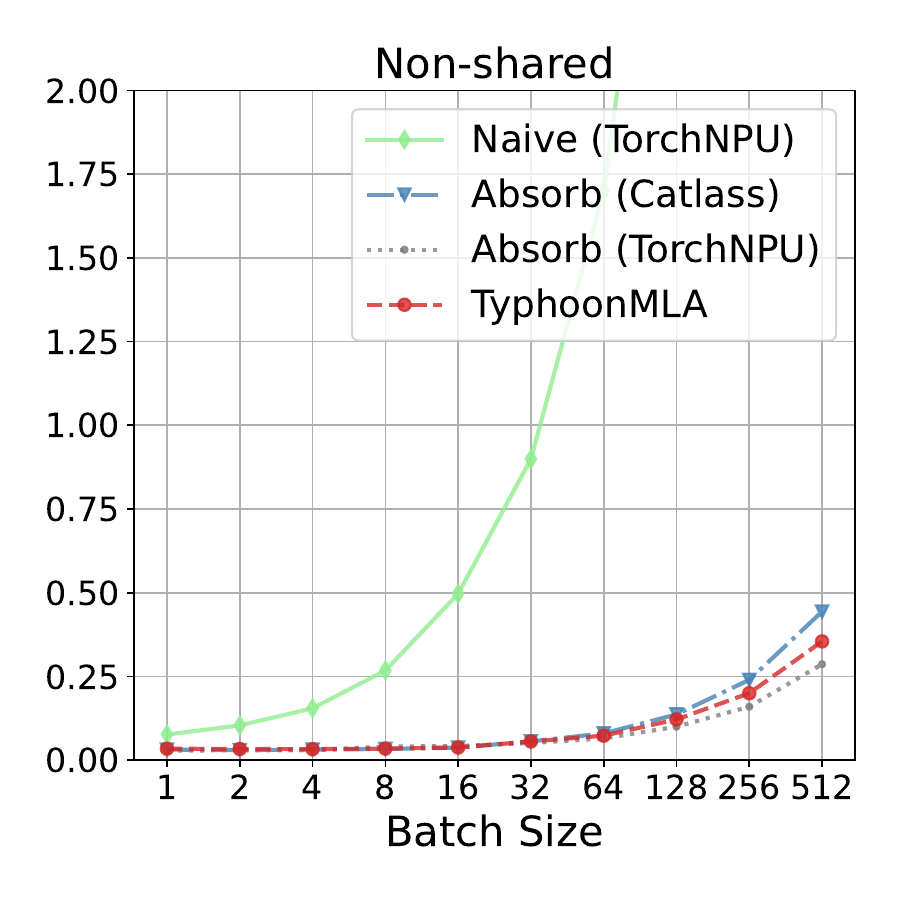}}
\subfigure[Full execution]{\label{fig:benchmark-full}\includegraphics[width=0.3\textwidth]{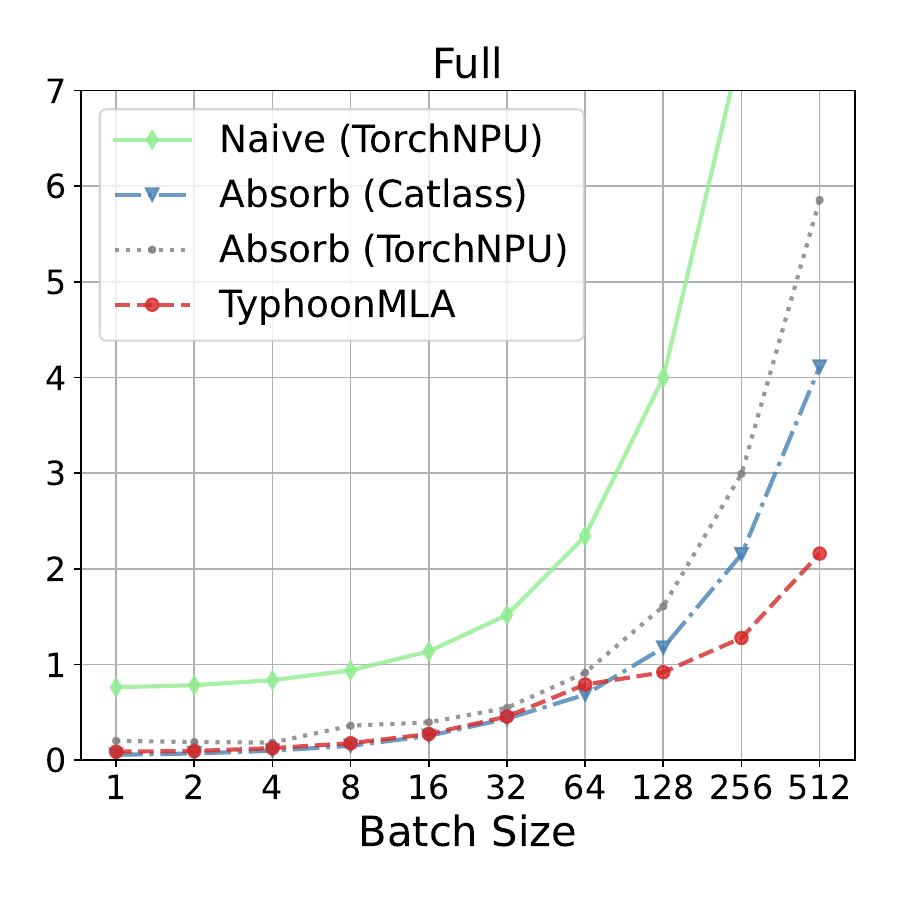}}
\caption{Performance breakdown of TyphoonMLA and the baselines on Ascend NPUs for DeepSeek-v3, assuming a shared prefix of length 4096 and a query length of 128. Execution time of individual components is measured using the CANN toolkit's msprof tool.}
\label{fig:benchmark}
\end{figure}

\subsection{Sensitivity to batch size}
To analyze how different components of TyphoonMLA benefit from data reuse, we profile TyphoonMLA and the baseline methods on an Ascend NPU across a range of batch sizes during the execution of DeepSeek-v3. Fig.\ref{fig:benchmark} shows the execution time of the shared part, non-shared part, and the overall attention calculation. Consistent with the complexity analysis in Section \ref{sec:complexity}, the absorb baseline outperforms the naive baseline at small batch sizes in the shared part of the attention calculations due to the reduced memory bandwidth requirements of the compressed KV-cache in latent space, as shown in Fig.\ref{fig:benchmark-stage1}. As the batch size increases, the execution time of the absorb baselines grows nearly linearly, while the naive baseline remains about the same, thanks to improved operational intensity. Around a batch size of 64, the naive baseline becomes faster than the absorb. To maintain optimal performance across all batch sizes, TyphoonMLA switches from the absorb-only implementation to the mixed naive-absorb kernel once the batch size exceeds this cut-off point.  

Fig.\ref{fig:benchmark-stage2} shows the execution time of TyphoonMLA and the baseline methods for the non-shared part of the attention calculations. Since this part does not contain any shared prefix, increasing the batch size does not improve the operational intensity; hence, the absorb implementations remain faster than the naive across all batch sizes. Consequently, TyphoonMLA consistently employs the absorb implementation for the non-shared part. When these two portions are combined, as shown in Fig.\ref{fig:benchmark-full}, TyphoonMLA behaves identically to the absorb baseline up to the cut-off batch size of 64, beyond which TyphoonMLA outperforms the absorb baseline, achieving a speedup of up to 2$\times$ at the batch size of 512. These results confirm that TyphoonMLA’s performance gains primarily arise from exploiting the shared portion of the KV-cache.

%% file: biblio.bib
@article{OpenAI23,
  author       = {OpenAI},
  title        = {{GPT-4} Technical Report},
  journal      = {CoRR},
  volume       = {abs/2303.08774},
  year         = {2023},
  url          = {https://doi.org/10.48550/arXiv.2303.08774},
  doi          = {10.48550/ARXIV.2303.08774},
  eprinttype    = {arXiv},
  eprint       = {2303.08774},
  bibsource    = {dblp computer science bibliography, https://dblp.org}
}

@article{Chen21,
  author       = {Mark Chen and
                  Jerry Tworek and
                  Heewoo Jun and
                  Qiming Yuan and
                  Henrique Pond{\'{e}} de Oliveira Pinto and
                  Jared Kaplan and
                  Harri Edwards and
                  Yuri Burda and
                  Nicholas Joseph and
                  Greg Brockman and
                  Alex Ray and
                  Raul Puri and
                  Gretchen Krueger and
                  Michael Petrov and
                  Heidy Khlaaf and
                  Girish Sastry and
                  Pamela Mishkin and
                  Brooke Chan and
                  Scott Gray and
                  Nick Ryder and
                  Mikhail Pavlov and
                  Alethea Power and
                  Lukasz Kaiser and
                  Mohammad Bavarian and
                  Clemens Winter and
                  Philippe Tillet and
                  Felipe Petroski Such and
                  Dave Cummings and
                  Matthias Plappert and
                  Fotios Chantzis and
                  Elizabeth Barnes and
                  Ariel Herbert{-}Voss and
                  William Hebgen Guss and
                  Alex Nichol and
                  Alex Paino and
                  Nikolas Tezak and
                  Jie Tang and
                  Igor Babuschkin and
                  Suchir Balaji and
                  Shantanu Jain and
                  William Saunders and
                  Christopher Hesse and
                  Andrew N. Carr and
                  Jan Leike and
                  Joshua Achiam and
                  Vedant Misra and
                  Evan Morikawa and
                  Alec Radford and
                  Matthew Knight and
                  Miles Brundage and
                  Mira Murati and
                  Katie Mayer and
                  Peter Welinder and
                  Bob McGrew and
                  Dario Amodei and
                  Sam McCandlish and
                  Ilya Sutskever and
                  Wojciech Zaremba},
  title        = {Evaluating Large Language Models Trained on Code},
  journal      = {CoRR},
  volume       = {abs/2107.03374},
  year         = {2021},
  url          = {https://arxiv.org/abs/2107.03374},
  eprinttype   = {arXiv},
  eprint       = {2107.03374},
  pages        = {},
  bibsource    = {dblp computer science bibliography, https://dblp.org}
}

@article{DeepSeek2024,
  author       = {DeepSeek{-}AI and
                  Aixin Liu and
                  Bei Feng and
                  Bin Wang and
                  Bingxuan Wang and
                  Bo Liu and
                  Chenggang Zhao and
                  Chengqi Deng and
                  Chong Ruan and
                  Damai Dai and
                  Daya Guo and
                  Dejian Yang and
                  Deli Chen and
                  Dongjie Ji and
                  Erhang Li and
                  Fangyun Lin and
                  Fuli Luo and
                  Guangbo Hao and
                  Guanting Chen and
                  Guowei Li and
                  Hao Zhang and
                  Hanwei Xu and
                  Hao Yang and
                  Haowei Zhang and
                  Honghui Ding and
                  Huajian Xin and
                  Huazuo Gao and
                  Hui Li and
                  Hui Qu and
                  J. L. Cai and
                  Jian Liang and
                  Jianzhong Guo and
                  Jiaqi Ni and
                  Jiashi Li and
                  Jin Chen and
                  Jingyang Yuan and
                  Junjie Qiu and
                  Junxiao Song and
                  Kai Dong and
                  Kaige Gao and
                  Kang Guan and
                  Lean Wang and
                  Lecong Zhang and
                  Lei Xu and
                  Leyi Xia and
                  Liang Zhao and
                  Liyue Zhang and
                  Meng Li and
                  Miaojun Wang and
                  Mingchuan Zhang and
                  Minghua Zhang and
                  Minghui Tang and
                  Mingming Li and
                  Ning Tian and
                  Panpan Huang and
                  Peiyi Wang and
                  Peng Zhang and
                  Qihao Zhu and
                  Qinyu Chen and
                  Qiushi Du and
                  R. J. Chen and
                  R. L. Jin and
                  Ruiqi Ge and
                  Ruizhe Pan and
                  Runxin Xu and
                  Ruyi Chen and
                  S. S. Li and
                  Shanghao Lu and
                  Shangyan Zhou and
                  Shanhuang Chen and
                  Shaoqing Wu and
                  Shengfeng Ye and
                  Shirong Ma and
                  Shiyu Wang and
                  Shuang Zhou and
                  Shuiping Yu and
                  Shunfeng Zhou and
                  Size Zheng and
                  Tao Wang and
                  Tian Pei and
                  Tian Yuan and
                  Tianyu Sun and
                  W. L. Xiao and
                  Wangding Zeng and
                  Wei An and
                  Wen Liu and
                  Wenfeng Liang and
                  Wenjun Gao and
                  Wentao Zhang and
                  X. Q. Li and
                  Xiangyue Jin and
                  Xianzu Wang and
                  Xiao Bi and
                  Xiaodong Liu and
                  Xiaohan Wang and
                  Xiaojin Shen and
                  Xiaokang Chen and
                  Xiaosha Chen and
                  Xiaotao Nie and
                  Xiaowen Sun and
                  Zihan Wang and
                  et al.},
  title        = {DeepSeek-V2: {A} Strong, Economical, and Efficient Mixture-of-Experts
                  Language Model},
  journal      = {CoRR},
  volume       = {abs/2405.04434},
  year         = {2024},
  url          = {https://doi.org/10.48550/arXiv.2405.04434},
  doi          = {10.48550/ARXIV.2405.04434},
  eprinttype    = {arXiv},
  eprint       = {2405.04434},
  bibsource    = {dblp computer science bibliography, https://dblp.org}
}

@inproceedings{Vaswani17,
  author       = {Ashish Vaswani and
                  Noam Shazeer and
                  Niki Parmar and
                  Jakob Uszkoreit and
                  Llion Jones and
                  Aidan N. Gomez and
                  Lukasz Kaiser and
                  Illia Polosukhin},
  editor       = {Isabelle Guyon and
                  Ulrike von Luxburg and
                  Samy Bengio and
                  Hanna M. Wallach and
                  Rob Fergus and
                  S. V. N. Vishwanathan and
                  Roman Garnett},
  title        = {Attention is All you Need},
  booktitle    = {Advances in Neural Information Processing Systems 30: Annual Conference
                  on Neural Information Processing Systems 2017, December 4-9, 2017,
                  Long Beach, CA, {USA}},
  pages        = {5998--6008},
  year         = {2017},
  url          = {https://proceedings.neurips.cc/paper/2017/hash/3f5ee243547dee91fbd053c1c4a845aa-Abstract.html},
  bibsource    = {dblp computer science bibliography, https://dblp.org}
}

@inproceedings{Ainslie23,
  author       = {Joshua Ainslie and
                  James Lee{-}Thorp and
                  Michiel de Jong and
                  Yury Zemlyanskiy and
                  Federico Lebr{\'{o}}n and
                  Sumit Sanghai},
  editor       = {Houda Bouamor and
                  Juan Pino and
                  Kalika Bali},
  title        = {{GQA:} Training Generalized Multi-Query Transformer Models from Multi-Head
                  Checkpoints},
  booktitle    = {Proceedings of the 2023 Conference on Empirical Methods in Natural
                  Language Processing, {EMNLP} 2023, Singapore, December 6-10, 2023},
  pages        = {4895--4901},
  publisher    = {Association for Computational Linguistics},
  year         = {2023},
  url          = {https://doi.org/10.18653/v1/2023.emnlp-main.298},
  doi          = {10.18653/V1/2023.EMNLP-MAIN.298},
  bibsource    = {dblp computer science bibliography, https://dblp.org}
}

@article{Juravsky24,
  author       = {Jordan Juravsky and
                  Bradley C. A. Brown and
                  Ryan Ehrlich and
                  Daniel Y. Fu and
                  Christopher R{\'{e}} and
                  Azalia Mirhoseini},
  title        = {Hydragen: High-Throughput {LLM} Inference with Shared Prefixes},
  journal      = {CoRR},
  volume       = {abs/2402.05099},
  year         = {2024},
  url          = {https://doi.org/10.48550/arXiv.2402.05099},
  doi          = {10.48550/ARXIV.2402.05099},
  eprinttype    = {arXiv},
  eprint       = {2402.05099},
  bibsource    = {dblp computer science bibliography, https://dblp.org}
}

@misc{system_prompt_repo,
  author = {Á. Johnson},
  title = {{System Prompts Leaks}},
  year = 2025,
  url = {https://github.com/asgeirtj/system_prompts_leaks},
  urldate = {2025/09/09}
}

@misc{deepseek-profile,
  author = {Deepseek-AI},
  title = {{Profiling Data in DeepSeek Infra}},
  year = 2025,
  url = {https://github.com/deepseek-ai/profile-data},
  urldate = {2025/11/26}
}

@inproceedings{Yao23,
  author       = {Shunyu Yao and
                  Dian Yu and
                  Jeffrey Zhao and
                  Izhak Shafran and
                  Tom Griffiths and
                  Yuan Cao and
                  Karthik Narasimhan},
  editor       = {Alice Oh and
                  Tristan Naumann and
                  Amir Globerson and
                  Kate Saenko and
                  Moritz Hardt and
                  Sergey Levine},
  title        = {Tree of Thoughts: Deliberate Problem Solving with Large Language Models},
  booktitle    = {Advances in Neural Information Processing Systems 36: Annual Conference
                  on Neural Information Processing Systems 2023, NeurIPS 2023, New Orleans,
                  LA, USA, December 10 - 16, 2023},
  year         = {2023},
  url          = {http://papers.nips.cc/paper\_files/paper/2023/hash/271db9922b8d1f4dd7aaef84ed5ac703-Abstract-Conference.html},
  bibsource    = {dblp computer science bibliography, https://dblp.org}
}

@inproceedings{Besta24,
  author       = {Maciej Besta and
                  Nils Blach and
                  Ales Kubicek and
                  Robert Gerstenberger and
                  Michal Podstawski and
                  Lukas Gianinazzi and
                  Joanna Gajda and
                  Tomasz Lehmann and
                  Hubert Niewiadomski and
                  Piotr Nyczyk and
                  Torsten Hoefler},
  editor       = {Michael J. Wooldridge and
                  Jennifer G. Dy and
                  Sriraam Natarajan},
  title        = {Graph of Thoughts: Solving Elaborate Problems with Large Language
                  Models},
  booktitle    = {Thirty-Eighth {AAAI} Conference on Artificial Intelligence, {AAAI}
                  2024, Thirty-Sixth Conference on Innovative Applications of Artificial
                  Intelligence, {IAAI} 2024, Fourteenth Symposium on Educational Advances
                  in Artificial Intelligence, {EAAI} 2014, February 20-27, 2024, Vancouver,
                  Canada},
  pages        = {17682--17690},
  publisher    = {{AAAI} Press},
  year         = {2024},
  url          = {https://doi.org/10.1609/aaai.v38i16.29720},
  doi          = {10.1609/AAAI.V38I16.29720},
  bibsource    = {dblp computer science bibliography, https://dblp.org}
}

@article{Wang25-opt,
  author       = {Jikai Wang and
                  Yi Su and
                  Juntao Li and
                  Qingrong Xia and
                  Zi Ye and
                  Xinyu Duan and
                  Zhefeng Wang and
                  Min Zhang},
  title        = {OPT-Tree: Speculative Decoding with Adaptive Draft Tree Structure},
  journal      = {Trans. Assoc. Comput. Linguistics},
  volume       = {13},
  pages        = {188--199},
  year         = {2025},
  url          = {https://doi.org/10.1162/tacl\_a\_00735},
  doi          = {10.1162/TACL\_A\_00735},
  bibsource    = {dblp computer science bibliography, https://dblp.org}
}

@inproceedings{Yao25,
  author       = {Jinwei Yao and
                  Kaiqi Chen and
                  Kexun Zhang and
                  Jiaxuan You and
                  Binhang Yuan and
                  Zeke Wang and
                  Tao Lin},
  title        = {DeFT: Decoding with Flash Tree-attention for Efficient Tree-structured
                  {LLM} Inference},
  booktitle    = {The Thirteenth International Conference on Learning Representations,
                  {ICLR} 2025, Singapore, April 24-28, 2025},
  publisher    = {OpenReview.net},
  year         = {2025},
  url          = {https://openreview.net/forum?id=2c7pfOqu9k},
  bibsource    = {dblp computer science bibliography, https://dblp.org}
}

@inproceedings{Pan2025,
  title={FastTree: Optimizing Attention Kernel and Runtime for Tree-Structured LLM Inference},
  author={Pan, Zaifeng and Ding, Yitong and Guan, Yue and Wang, Zheng and Yu, Zhongkai and Tang, Xulong and Wang, Yida and Ding, Yufei},
  booktitle={Eighth Conference on Machine Learning and Systems},
  year={2025}
}

@article{Wang25-flash,
  author       = {Zhibin Wang and
                  Rui Ning and
                  Chao Fang and
                  Zhonghui Zhang and
                  Xi Lin and
                  Shaobo Ma and
                  Mo Zhou and
                  Xue Li and
                  Zhongfeng Wang and
                  Chengying Huan and
                  Rong Gu and
                  Kun Yang and
                  Guihai Chen and
                  Sheng Zhong and
                  Chen Tian},
  title        = {FlashForge: Ultra-Efficient Prefix-Aware Attention for {LLM} Decoding},
  journal      = {CoRR},
  volume       = {abs/2505.17694},
  year         = {2025},
  url          = {https://doi.org/10.48550/arXiv.2505.17694},
  doi          = {10.48550/ARXIV.2505.17694},
  eprinttype    = {arXiv},
  eprint       = {2505.17694},
  bibsource    = {dblp computer science bibliography, https://dblp.org}
}

@article{deepseekv3,
  author       = {DeepSeek{-}AI and
                  Aixin Liu and
                  Bei Feng and
                  Bing Xue and
                  Bingxuan Wang and
                  Bochao Wu and
                  Chengda Lu and
                  Chenggang Zhao and
                  Chengqi Deng and
                  Chenyu Zhang and
                  Chong Ruan and
                  Damai Dai and
                  Daya Guo and
                  Dejian Yang and
                  Deli Chen and
                  Dongjie Ji and
                  Erhang Li and
                  Fangyun Lin and
                  Fucong Dai and
                  Fuli Luo and
                  Guangbo Hao and
                  Guanting Chen and
                  Guowei Li and
                  H. Zhang and
                  Han Bao and
                  Hanwei Xu and
                  Haocheng Wang and
                  Haowei Zhang and
                  Honghui Ding and
                  Huajian Xin and
                  Huazuo Gao and
                  Hui Li and
                  Hui Qu and
                  J. L. Cai and
                  Jian Liang and
                  Jianzhong Guo and
                  Jiaqi Ni and
                  Jiashi Li and
                  Jiawei Wang and
                  Jin Chen and
                  Jingchang Chen and
                  Jingyang Yuan and
                  Junjie Qiu and
                  Junlong Li and
                  Junxiao Song and
                  Kai Dong and
                  Kai Hu and
                  Kaige Gao and
                  Kang Guan and
                  Kexin Huang and
                  Kuai Yu and
                  Lean Wang and
                  Lecong Zhang and
                  Lei Xu and
                  Leyi Xia and
                  Liang Zhao and
                  Litong Wang and
                  Liyue Zhang and
                  Meng Li and
                  Miaojun Wang and
                  Mingchuan Zhang and
                  Minghua Zhang and
                  Minghui Tang and
                  Mingming Li and
                  Ning Tian and
                  Panpan Huang and
                  Peiyi Wang and
                  Peng Zhang and
                  Qiancheng Wang and
                  Qihao Zhu and
                  Qinyu Chen and
                  Qiushi Du and
                  R. J. Chen and
                  R. L. Jin and
                  Ruiqi Ge and
                  Ruisong Zhang and
                  Ruizhe Pan and
                  Runji Wang and
                  Runxin Xu and
                  Ruoyu Zhang and
                  Ruyi Chen and
                  S. S. Li and
                  Shanghao Lu and
                  Shangyan Zhou and
                  Shanhuang Chen and
                  Shaoqing Wu and
                  Shengfeng Ye and
                  Shirong Ma and
                  Shiyu Wang and
                  Shuang Zhou and
                  Shuiping Yu and
                  Shunfeng Zhou and
                  Shuting Pan and
                  T. Wang and
                  Tao Yun and
                  Tian Pei and
                  Tianyu Sun and
                  W. L. Xiao and
                  Wangding Zeng},
  title        = {DeepSeek-V3 Technical Report},
  journal      = {CoRR},
  volume       = {abs/2412.19437},
  year         = {2024},
  url          = {https://doi.org/10.48550/arXiv.2412.19437},
  doi          = {10.48550/ARXIV.2412.19437},
  eprinttype    = {arXiv},
  eprint       = {2412.19437},
  bibsource    = {dblp computer science bibliography, https://dblp.org}
}

@article{kimik2,
  author       = {Yifan Bai and
                  Yiping Bao and
                  Guanduo Chen and
                  Jiahao Chen and
                  Ningxin Chen and
                  Ruijue Chen and
                  Yanru Chen and
                  Yuankun Chen and
                  Yutian Chen and
                  Zhuofu Chen and
                  Jialei Cui and
                  Hao Ding and
                  Mengnan Dong and
                  Angang Du and
                  Chenzhuang Du and
                  Dikang Du and
                  Yulun Du and
                  Yu Fan and
                  Yichen Feng and
                  Kelin Fu and
                  Bofei Gao and
                  Hongcheng Gao and
                  Peizhong Gao and
                  Tong Gao and
                  Xinran Gu and
                  Longyu Guan and
                  Haiqing Guo and
                  Jianhang Guo and
                  Hao Hu and
                  Xiaoru Hao and
                  Tianhong He and
                  Weiran He and
                  Wenyang He and
                  Chao Hong and
                  Yangyang Hu and
                  Zhenxing Hu and
                  Weixiao Huang and
                  Zhiqi Huang and
                  Zihao Huang and
                  Tao Jiang and
                  Zhejun Jiang and
                  Xinyi Jin and
                  Yongsheng Kang and
                  Guokun Lai and
                  Cheng Li and
                  Fang Li and
                  Haoyang Li and
                  Ming Li and
                  Wentao Li and
                  Yanhao Li and
                  Yiwei Li and
                  Zhaowei Li and
                  Zheming Li and
                  Hongzhan Lin and
                  Xiaohan Lin and
                  Zongyu Lin and
                  Chengyin Liu and
                  Chenyu Liu and
                  Hongzhang Liu and
                  Jingyuan Liu and
                  Junqi Liu and
                  Liang Liu and
                  Shaowei Liu and
                  T. Y. Liu and
                  Tianwei Liu and
                  Weizhou Liu and
                  Yangyang Liu and
                  Yibo Liu and
                  Yiping Liu and
                  Yue Liu and
                  Zhengying Liu and
                  Enzhe Lu and
                  Lijun Lu and
                  Shengling Ma and
                  Xinyu Ma and
                  Yingwei Ma and
                  Shaoguang Mao and
                  Jie Mei and
                  Xin Men and
                  Yibo Miao and
                  Siyuan Pan and
                  Yebo Peng and
                  Ruoyu Qin and
                  Bowen Qu and
                  Zeyu Shang and
                  Lidong Shi and
                  Shengyuan Shi and
                  Feifan Song and
                  Jianlin Su and
                  Zhengyuan Su and
                  Xinjie Sun and
                  Flood Sung and
                  Heyi Tang and
                  Jiawen Tao and
                  Qifeng Teng and
                  Chensi Wang and
                  Dinglu Wang and
                  Feng Wang and
                  Haiming Wang},
  title        = {Kimi {K2:} Open Agentic Intelligence},
  journal      = {CoRR},
  volume       = {abs/2507.20534},
  year         = {2025},
  url          = {https://doi.org/10.48550/arXiv.2507.20534},
  doi          = {10.48550/ARXIV.2507.20534},
  eprinttype    = {arXiv},
  eprint       = {2507.20534},
  bibsource    = {dblp computer science bibliography, https://dblp.org}
}

@inproceedings{Dao22,
  author       = {Tri Dao and
                  Daniel Y. Fu and
                  Stefano Ermon and
                  Atri Rudra and
                  Christopher R{\'{e}}},
  editor       = {Sanmi Koyejo and
                  S. Mohamed and
                  A. Agarwal and
                  Danielle Belgrave and
                  K. Cho and
                  A. Oh},
  title        = {FlashAttention: Fast and Memory-Efficient Exact Attention with IO-Awareness},
  booktitle    = {Advances in Neural Information Processing Systems 35: Annual Conference
                  on Neural Information Processing Systems 2022, NeurIPS 2022, New Orleans,
                  LA, USA, November 28 - December 9, 2022},
  year         = {2022},
  url          = {http://papers.nips.cc/paper\_files/paper/2022/hash/67d57c32e20fd0a7a302cb81d36e40d5-Abstract-Conference.html},
  bibsource    = {dblp computer science bibliography, https://dblp.org}
}

@misc{flashmla2025,
      title={FlashMLA: Efficient MLA decoding kernels},
      author={Jiashi Li, Shengyu Liu},
      year={2025},
      publisher = {GitHub},
      howpublished = {\url{https://github.com/deepseek-ai/FlashMLA}},
}

@inproceedings{Wang23,
  author       = {Xuezhi Wang and
                  Jason Wei and
                  Dale Schuurmans and
                  Quoc V. Le and
                  Ed H. Chi and
                  Sharan Narang and
                  Aakanksha Chowdhery and
                  Denny Zhou},
  title        = {Self-Consistency Improves Chain of Thought Reasoning in Language Models},
  booktitle    = {The Eleventh International Conference on Learning Representations,
                  {ICLR} 2023, Kigali, Rwanda, May 1-5, 2023},
  publisher    = {OpenReview.net},
  year         = {2023},
  url          = {https://openreview.net/forum?id=1PL1NIMMrw},
  bibsource    = {dblp computer science bibliography, https://dblp.org}
}

@article{Williams09,
  author       = {Samuel Williams and
                  Andrew Waterman and
                  David A. Patterson},
  title        = {Roofline: an insightful visual performance model for multicore architectures},
  journal      = {Commun. {ACM}},
  volume       = {52},
  number       = {4},
  pages        = {65--76},
  year         = {2009},
  url          = {https://doi.org/10.1145/1498765.1498785},
  doi          = {10.1145/1498765.1498785},
  bibsource    = {dblp computer science bibliography, https://dblp.org}
}

@misc{torchnpu,
  author = {Huawei},
  title = {{Ascend Extension for PyTorch}},
  year = 2024,
  url = {https://gitee.com/ascend/pytorch},
}

@misc{cann,
  author = {Huawei},
  title = {{CANN Community Edition}},
  year = 2024,
  url = {https://www.hiascend.com/en/software/cann/community},
  urldate = {2024/01/05}
}

@misc{catlass,
  author = {Huawei},
  title = {{Ascend CATLASS}},
  year = 2025,
  url = {https://gitee.com/ascend/catlass},
}

@article{ye25,
  title={Flashinfer: Efficient and customizable attention engine for llm inference serving},
  author={Ye, Zihao and Chen, Lequn and Lai, Ruihang and Lin, Wuwei and Zhang, Yineng and Wang, Stephanie and Chen, Tianqi and Kasikci, Baris and Grover, Vinod and Krishnamurthy, Arvind and others},
  journal={arXiv preprint arXiv:2501.01005},
  year={2025}
}

@inproceedings{Hendrycks21,
  author       = {Dan Hendrycks and
                  Collin Burns and
                  Steven Basart and
                  Andy Zou and
                  Mantas Mazeika and
                  Dawn Song and
                  Jacob Steinhardt},
  title        = {Measuring Massive Multitask Language Understanding},
  booktitle    = {9th International Conference on Learning Representations, {ICLR} 2021,
                  Virtual Event, Austria, May 3-7, 2021},
  publisher    = {OpenReview.net},
  year         = {2021},
  url          = {https://openreview.net/forum?id=d7KBjmI3GmQ},
  bibsource    = {dblp computer science bibliography, https://dblp.org}
}

@article{Cobbe21,
  author       = {Karl Cobbe and
                  Vineet Kosaraju and
                  Mohammad Bavarian and
                  Mark Chen and
                  Heewoo Jun and
                  Lukasz Kaiser and
                  Matthias Plappert and
                  Jerry Tworek and
                  Jacob Hilton and
                  Reiichiro Nakano and
                  Christopher Hesse and
                  John Schulman},
  title        = {Training Verifiers to Solve Math Word Problems},
  journal      = {CoRR},
  volume       = {abs/2110.14168},
  year         = {2021},
  url          = {https://arxiv.org/abs/2110.14168},
  eprinttype    = {arXiv},
  eprint       = {2110.14168},
  bibsource    = {dblp computer science bibliography, https://dblp.org}
}

@article{Wei24,
  author       = {Jason Wei and
                  Nguyen Karina and
                  Hyung Won Chung and
                  Yunxin Joy Jiao and
                  Spencer Papay and
                  Amelia Glaese and
                  John Schulman and
                  William Fedus},
  title        = {Measuring short-form factuality in large language models},
  journal      = {CoRR},
  volume       = {abs/2411.04368},
  year         = {2024},
  url          = {https://doi.org/10.48550/arXiv.2411.04368},
  doi          = {10.48550/ARXIV.2411.04368},
  eprinttype    = {arXiv},
  eprint       = {2411.04368},
  bibsource    = {dblp computer science bibliography, https://dblp.org}
}

@inproceedings{dao23,
  title={Flash{A}ttention-2: Faster Attention with Better Parallelism and Work Partitioning},
  author={Dao, Tri},
  booktitle={International Conference on Learning Representations (ICLR)},
  year={2024}
}

@article{Dege25,
  author       = {Pengcuo Dege and
                  Qiuming Luo and
                  Rui Mao and
                  Chang Kong},
  title        = {FlashMLA-ETAP: Efficient Transpose Attention Pipeline for Accelerating
                  {MLA} Inference on {NVIDIA} {H20} GPUs},
  journal      = {CoRR},
  volume       = {abs/2506.01969},
  year         = {2025},
  url          = {https://doi.org/10.48550/arXiv.2506.01969},
  doi          = {10.48550/ARXIV.2506.01969},
  eprinttype    = {arXiv},
  eprint       = {2506.01969},
  bibsource    = {dblp computer science bibliography, https://dblp.org}
}

@inproceedings{Yu22,
  author       = {Gyeong{-}In Yu and
                  Joo Seong Jeong and
                  Geon{-}Woo Kim and
                  Soojeong Kim and
                  Byung{-}Gon Chun},
  editor       = {Marcos K. Aguilera and
                  Hakim Weatherspoon},
  title        = {Orca: {A} Distributed Serving System for Transformer-Based Generative
                  Models},
  booktitle    = {16th {USENIX} Symposium on Operating Systems Design and Implementation,
                  {OSDI} 2022, Carlsbad, CA, USA, July 11-13, 2022},
  pages        = {521--538},
  publisher    = {{USENIX} Association},
  year         = {2022},
  url          = {https://www.usenix.org/conference/osdi22/presentation/yu},
  bibsource    = {dblp computer science bibliography, https://dblp.org}
}

@inproceedings{Kwon23,
  author       = {Woosuk Kwon and
                  Zhuohan Li and
                  Siyuan Zhuang and
                  Ying Sheng and
                  Lianmin Zheng and
                  Cody Hao Yu and
                  Joseph Gonzalez and
                  Hao Zhang and
                  Ion Stoica},
  editor       = {Jason Flinn and
                  Margo I. Seltzer and
                  Peter Druschel and
                  Antoine Kaufmann and
                  Jonathan Mace},
  title        = {Efficient Memory Management for Large Language Model Serving with
                  PagedAttention},
  booktitle    = {Proceedings of the 29th Symposium on Operating Systems Principles,
                  {SOSP} 2023, Koblenz, Germany, October 23-26, 2023},
  pages        = {611--626},
  publisher    = {{ACM}},
  year         = {2023},
  url          = {https://doi.org/10.1145/3600006.3613165},
  doi          = {10.1145/3600006.3613165},
  bibsource    = {dblp computer science bibliography, https://dblp.org}
}

@article{Shoeybi19,
  author       = {Mohammad Shoeybi and
                  Mostofa Patwary and
                  Raul Puri and
                  Patrick LeGresley and
                  Jared Casper and
                  Bryan Catanzaro},
  title        = {Megatron-LM: Training Multi-Billion Parameter Language Models Using
                  Model Parallelism},
  journal      = {CoRR},
  volume       = {abs/1909.08053},
  year         = {2019},
  url          = {http://arxiv.org/abs/1909.08053},
  eprinttype    = {arXiv},
  eprint       = {1909.08053},
  bibsource    = {dblp computer science bibliography, https://dblp.org}
}

@article{Jacobs23,
  author       = {Sam Ade Jacobs and
                  Masahiro Tanaka and
                  Chengming Zhang and
                  Minjia Zhang and
                  Shuaiwen Leon Song and
                  Samyam Rajbhandari and
                  Yuxiong He},
  title        = {DeepSpeed Ulysses: System Optimizations for Enabling Training of Extreme
                  Long Sequence Transformer Models},
  journal      = {CoRR},
  volume       = {abs/2309.14509},
  year         = {2023},
  url          = {https://doi.org/10.48550/arXiv.2309.14509},
  doi          = {10.48550/ARXIV.2309.14509},
  eprinttype    = {arXiv},
  eprint       = {2309.14509},
  bibsource    = {dblp computer science bibliography, https://dblp.org}
}

@inproceedings{Zheng24,
  author       = {Lianmin Zheng and
                  Liangsheng Yin and
                  Zhiqiang Xie and
                  Chuyue Sun and
                  Jeff Huang and
                  Cody Hao Yu and
                  Shiyi Cao and
                  Christos Kozyrakis and
                  Ion Stoica and
                  Joseph E. Gonzalez and
                  Clark W. Barrett and
                  Ying Sheng},
  editor       = {Amir Globersons and
                  Lester Mackey and
                  Danielle Belgrave and
                  Angela Fan and
                  Ulrich Paquet and
                  Jakub M. Tomczak and
                  Cheng Zhang},
  title        = {SGLang: Efficient Execution of Structured Language Model Programs},
  booktitle    = {Advances in Neural Information Processing Systems 38: Annual Conference
                  on Neural Information Processing Systems 2024, NeurIPS 2024, Vancouver,
                  BC, Canada, December 10 - 15, 2024},
  year         = {2024},
  url          = {http://papers.nips.cc/paper\_files/paper/2024/hash/724be4472168f31ba1c9ac630f15dec8-Abstract-Conference.html},
  bibsource    = {dblp computer science bibliography, https://dblp.org}
}

@inproceedings{Ye24,
  author       = {Lu Ye and
                  Ze Tao and
                  Yong Huang and
                  Yang Li},
  editor       = {Lun{-}Wei Ku and
                  Andre Martins and
                  Vivek Srikumar},
  title        = {ChunkAttention: Efficient Self-Attention with Prefix-Aware {KV} Cache
                  and Two-Phase Partition},
  booktitle    = {Proceedings of the 62nd Annual Meeting of the Association for Computational
                  Linguistics (Volume 1: Long Papers), {ACL} 2024, Bangkok, Thailand,
                  August 11-16, 2024},
  pages        = {11608--11620},
  publisher    = {Association for Computational Linguistics},
  year         = {2024},
  url          = {https://doi.org/10.18653/v1/2024.acl-long.623},
  doi          = {10.18653/V1/2024.ACL-LONG.623},
  bibsource    = {dblp computer science bibliography, https://dblp.org}
}

@inproceedings{Zhu24,
  author       = {Lei Zhu and
                  Xinjiang Wang and
                  Wayne Zhang and
                  Rynson W. H. Lau},
  editor       = {Lun{-}Wei Ku and
                  Andre Martins and
                  Vivek Srikumar},
  title        = {RelayAttention for Efficient Large Language Model Serving with Long
                  System Prompts},
  booktitle    = {Proceedings of the 62nd Annual Meeting of the Association for Computational
                  Linguistics (Volume 1: Long Papers), {ACL} 2024, Bangkok, Thailand,
                  August 11-16, 2024},
  pages        = {4945--4957},
  publisher    = {Association for Computational Linguistics},
  year         = {2024},
  url          = {https://doi.org/10.18653/v1/2024.acl-long.270},
  doi          = {10.18653/V1/2024.ACL-LONG.270},
  bibsource    = {dblp computer science bibliography, https://dblp.org}
}

@misc{Ye24cascade,
  title={Cascade inference: Memory bandwidth efficient shared prefix batch decoding},
  author={Ye, Zihao and Lai, Ruihang and Lu, Bo-Ru and Lin, Chien-Yu and Zheng, Size and Chen, Lequn and Chen, Tianqi and Ceze, Luis},
  year={2024},
  publisher={February}
}

@misc{Spector25,
  title={Thunder{MLA}: Flashmla, faster and fuseder!},
  author={Benjamin Spector and Aaryan Singhal and Dan Fu and Chris Ré},
  year={2025},
  url={https://hazyresearch.stanford.edu/blog/2025-03-04-thundermla}
}

@article{longcat25,
  title={LongCat-Flash Technical Report},
  author={Meituan and Li, Bei and Lei, Bingye and Wang, Bo and Rong, Bolin and Wang, Chao and Zhang, Chao and Gao, Chen and Zhang, Chen and Sun, Cheng and others},
  journal={arXiv preprint arXiv:2509.01322},
  year={2025}
}

@article{Zuo25,
  author       = {Pengfei Zuo and
                  Huimin Lin and
                  Junbo Deng and
                  Nan Zou and
                  Xingkun Yang and
                  Yingyu Diao and
                  Weifeng Gao and
                  Ke Xu and
                  Zhangyu Chen and
                  Shirui Lu and
                  Zhao Qiu and
                  Peiyang Li and
                  Xianyu Chang and
                  Zhengzhong Yu and
                  Fangzheng Miao and
                  Jia Zheng and
                  Ying Li and
                  Yuan Feng and
                  Bei Wang and
                  Zaijian Zong and
                  Mosong Zhou and
                  Wenli Zhou and
                  Houjiang Chen and
                  Xingyu Liao and
                  Yipeng Li and
                  Wenxiao Zhang and
                  Ping Zhu and
                  Yinggang Wang and
                  Chuanjie Xiao and
                  Depeng Liang and
                  Dong Cao and
                  Juncheng Liu and
                  Yongqiang Yang and
                  Xiaolong Bai and
                  Yi Li and
                  Huaguo Xie and
                  Huatao Wu and
                  Zhibin Yu and
                  Lv Chen and
                  Hu Liu and
                  Yujun Ding and
                  Haipei Zhu and
                  Jing Xia and
                  Yi Xiong and
                  Zhou Yu and
                  Heng Liao},
  title        = {Serving Large Language Models on Huawei CloudMatrix384},
  journal      = {CoRR},
  volume       = {abs/2506.12708},
  year         = {2025},
  url          = {https://doi.org/10.48550/arXiv.2506.12708},
  doi          = {10.48550/ARXIV.2506.12708},
  eprinttype    = {arXiv},
  eprint       = {2506.12708},
  bibsource    = {dblp computer science bibliography, https://dblp.org}
}
